\documentclass{article} % For LaTeX2e
\usepackage{iclr2027_conference,times}

\usepackage{amsmath,amsfonts,bm}

\def\eqref#1{equation~\ref{#1}}
\def\1{\bm{1}}

\def\vtheta{{\bm{\theta}}}

\def\vd{{\bm{d}}}

\def\vg{{\bm{g}}}
\def\vh{{\bm{h}}}

\def\vu{{\bm{u}}}

\def\vx{{\bm{x}}}

\def\vz{{\bm{z}}}

\def\mA{{\bm{A}}}

\def\mF{{\bm{F}}}

\def\mH{{\bm{H}}}
\def\mI{{\bm{I}}}

\def\mLambda{{\bm{\Lambda}}}

\DeclareMathAlphabet{\mathsfit}{\encodingdefault}{\sfdefault}{m}{sl}
\SetMathAlphabet{\mathsfit}{bold}{\encodingdefault}{\sfdefault}{bx}{n}

\def\sB{{\mathbb{B}}}

\def\sK{{\mathbb{K}}}

\def\sS{{\mathbb{S}}}
\def\sT{{\mathbb{T}}}
\def\sU{{\mathbb{U}}}

\newcommand{\R}{\mathbb{R}}

\providecommand{\mzero}{\bm{0}}
\providecommand{\vDelta}{\bm{\Delta}}
\providecommand{\vepsilon}{\bm{\epsilon}}
\providecommand{\tr}{\operatorname{tr}}
\providecommand{\Diag}{\operatorname{Diag}}

\usepackage[hyphens]{url}
\usepackage{graphicx} % DO NOT CHANGE THIS
\usepackage{natbib}  % DO NOT CHANGE THIS AND DO NOT ADD ANY OPTIONS TO IT
\usepackage{caption} % DO NOT CHANGE THIS AND DO NOT ADD ANY OPTIONS TO IT
\usepackage{algorithm}
\usepackage{algorithmic}
\usepackage{bm}
\usepackage{multirow}
\usepackage{booktabs}
\usepackage{tabularx}

\usepackage{amsmath,amssymb} 
\newtheorem{proposition}{Proposition}

\newcommand{\ExpRes}[2]{\ensuremath{{#1}_{\scriptscriptstyle (#2)}}}

\newcommand{\minisection}[1]{\vspace{0.0in} \noindent {\bf #1}}

\title{Beyond Parameter Protection: Fisher-Shaped EWC for Replay-Free Continual Learning}

\author{
Liying Wang\textsuperscript{1,2},
Simone Magistri\textsuperscript{3},
Joost van de Weijer\textsuperscript{1,2}
\\[5pt]
{\small
\textsuperscript{1}Department of Computer Science,
Universitat Aut\`onoma de Barcelona, Spain}
\\
{\small
\textsuperscript{2}Computer Vision Center, Barcelona, Spain}
\\
{\small
\textsuperscript{3}Media Integration and Communication Center,
University of Florence, Italy}
\\[4pt]
{\small
\texttt{liying@cvc.uab.cat},
\texttt{simone.magistri@unifi.it},
\texttt{joost@cvc.uab.es}}
}
\iclrfinalcopy % Uncomment for camera-ready version, but NOT for submission.
\begin{document}

\maketitle
% Remove the conference header.
\fancyhead{}
\renewcommand{\headrulewidth}{0pt}

\begin{abstract}

In replay-free continual learning, Elastic Weight Consolidation
(EWC) uses Fisher information matrix (FIM) to protect acquired knowledge by
penalizing changes to historically important parameters.
However, a local importance estimate at the retained solution does not in general characterize sensitivity after finite parameter displacement.
This motivates controlling Fisher sensitivity over a neighborhood of potential future updates.
We propose Fisher-Shaped EWC (FS-EWC), which extends the role of
FIM from parameter protection to solution formation
for future retention.
During task learning, FS-EWC minimizes the worst-case
empirical FIM trace within a prescribed neighborhood, explicitly
favoring solutions that are less sensitive to bounded future
displacement.
At task completion, trajectory-based model merging constructs the
retained solution from multiple optimization states rather than a
single terminal iterate, after which FIM is estimated
for subsequent consolidation.
Experiments across multiple benchmarks show that FS-EWC improves performance over EWC, while controlled studies
indicate a more favorable stability--plasticity balance and greater
tolerance to parameter drift.
These results suggest that effective knowledge retention requires both
forming solutions with future retention in mind and protecting
acquired knowledge.
\end{abstract}

\section{Introduction}
\label{sec:introduction}

% Continual learning (CL) aims to enable learning systems to acquire new
% knowledge over time without forgetting previously learned
% capabilities~\citep{Wang2024survey}.
% This ability is particularly important in real-world environments
% where data distributions and learning demands continuously evolve~\citep{Shaheen2022}.
% Continual learning (CL) aims to enable learning systems to acquire new
% knowledge over time while retaining previously learned capabilities,
% a requirement that is particularly important in real-world environments
% where data distributions and learning demands continuously
% evolve~\citep{Wang2024survey,Shaheen2022}.
Continual learning (CL) aims to acquire new knowledge over time while
retaining previously learned capabilities, which is essential in
real-world environments where data distributions and learning demands
continuously evolve~\citep{Wang2024survey,Shaheen2022}.
% Ideally, a continual learner should reach a solution that preserves
% previously acquired knowledge while remaining sufficiently adaptable
% to new tasks.
This challenge is particularly pronounced in replay-free CL, where
historical data are unavailable during subsequent training.
% A continual learner should therefore preserve acquired knowledge while
% remaining sufficiently adaptable to new tasks.
% This objective is particularly challenging in replay-free CL, where
% historical data are unavailable during subsequent training.
% Ideally, a continual learner should preserve previously acquired
% knowledge while remaining sufficiently adaptable to new tasks.
% Achieving this objective becomes particularly challenging in
% replay-free CL, where historical data are unavailable
% during subsequent training.
% Obtaining this goal becomes more challenging in replay-free CL, where historical data are not
% available during subsequent training~\citep{
% vandevenThreeTypesIncremental2022a,
% MasanaClassIncrementalLearningSurvey}.
Elastic Weight Consolidation (EWC)~\citep{EWC2017} is a representative
and widely used regularization baseline for mitigating forgetting in
this setting.
However, it still faces a persistent stability--plasticity trade-off
between retaining acquired knowledge and adapting to new tasks.

EWC constructs future protection from the learned solution and
its fisher information matrix (FIM), penalizing changes to parameters that are
important to previously learned tasks.
After each task, the resulting solution is retained as the reference
for subsequent learning, while FIM is consolidated
with historical parameter importance~\citep{
huszar2018note, schwarz2018progress}.
Future updates are penalized according to this inherited importance,
only allowing movement along directions assigned lower FIM
values.
However, this protection is based on a local characterization of parameter
importance. As subsequent learning moves the model away from the
learned solution, directions that initially appear less important can
become increasingly sensitive.
This motivates learning a solution whose surrounding region is less
susceptible to future parameter drift, rather than relying only on
constraints imposed during later task.

% Current-task Fisher information therefore informs future protection
% without serving as an explicit sensitivity criterion for selecting
% the solution from which it is estimated.
% Among solutions with comparable current-task loss and historical
% penalty, EWC consequently does not explicitly favor one whose
% task loss is less sensitive to future parameter drift.

% A major line of EWC research improves consolidation by refining
% how parameter importance is estimated and represented.
% Structured representations address information lost by diagonal
% approximations~\citep{liu2018rotate,martens2015kfac}, while online
% and trajectory-based methods refine how importance is accumulated
% across learning~\citep{huszar2018note,SI2017Zenke,
% chaudhry2018riemannian}.Studies of Fisher computation examine empirical and model-based
% estimates and the effects of numerical
% approximations~\citep{martens2020new,vandeven2025computation}.
% These developments improve the information used to characterize
% learned parameters for subsequent protection.
% However, improving how or where importance is evaluated does not
% by itself introduce an objective that actively reduces the
% current task's sensitivity to future updates.
% This leaves a complementary question concerning the solution
% itself, rather than only the statistics used to protect it.

% Existing extensions of EWC mainly strengthen how parameter importance
% is constructed and used for protection.
Existing extensions of EWC mainly strengthen how parameter importance is constructed and used, while
still treating the solution to be protected as given.
They do not explicitly shape the current solution so that its
surrounding neighborhood remains less sensitive to future parameter
drift.
 Online and trajectory-aware approaches refine when and how importance
is accumulated across learning~\citep{
huszar2018note,schwarz2018progress,SI2017Zenke,
chaudhry2018riemannian};
studies of FIM computation examine how the FIM is
computed~\citep{martens2020new,vandeven2025computation};
and structured approximations improve how this information is
represented beyond a diagonal form~\citep{
martens2015kfac,liu2018rotate}.
% These advances improve the information used for parameter protection,
% but they still largely treat the solution to be protected as given.
% % They do not explicitly shape the current solution so that its
% % neighborhood remains less sensitive to future parameter drift.

\begin{figure}

    \centering
     \includegraphics[width=0.85\linewidth]{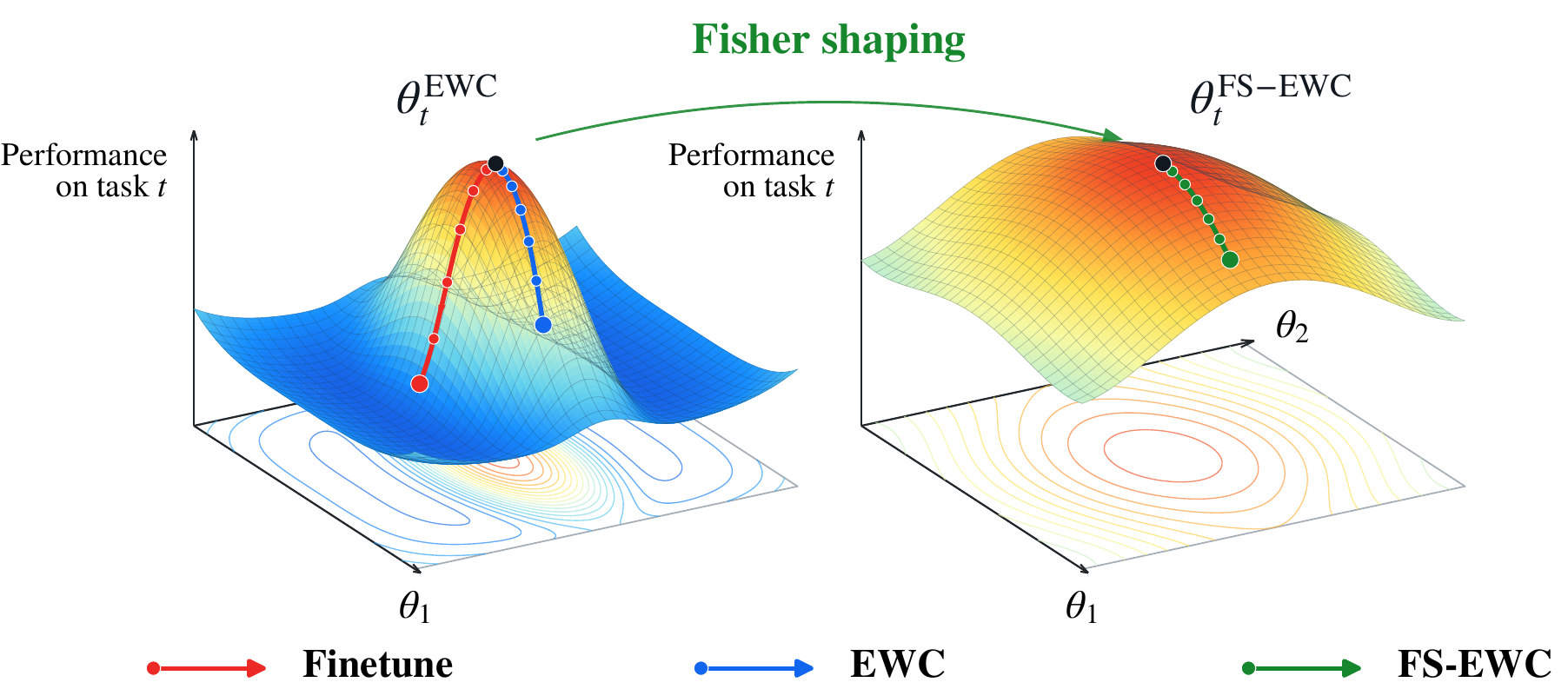}
     % {continual_learning_minima2_abc_narrative_v10.pdf}
     % {FS-EWC5.pdf}
     \caption{
     \textbf{Beyond protection: shaping solutions for future retention.}
% The surfaces schematically represent performance on task $t$, while
% the trajectories illustrate parameter updates induced by learning
% task $t+1$.
The surfaces schematically depict task-$t$ performance around learned
solutions, while trajectories illustrate parameter updates
during task-$t+1$ learning.
Unlike finetuning, EWC penalizes subsequent drift using
historical parameter importance.
FS-EWC additionally penalizes the worst-case empirical FIM
over a prescribed neighborhood during task-$t$ learning, favoring
solutions less sensitive to bounded future drift while preserving
historical protection. 
Complemented by model merging over task trajectory,
our approach incorporates future retention into how solutions are
learned and retained for subsequent protection.
}

    \label{fig:fs_ewc_overview}
\end{figure}

% Our analysis motivates using Fisher information not only to
% protect acquired knowledge, but also to guide the learning
% of solutions that better tolerate future parameter drift.
% Specifically, we bound task-specific empirical loss changes
% in terms of parameter drift and empirical Fisher information.
% For a prescribed drift radius, reducing neighborhood Fisher
% sensitivity tightens the upper bound on potential loss increases.
% This suggests complementing EWC's use of historical Fisher
% information for parameter protection with explicit optimization
% of current-task Fisher sensitivity while its data remain available.
% Future retention can be considered during current-task
% learning, rather than relying solely on protection in subsequent tasks.

The missing component is an explicit criterion for forming solutions
that remain less sensitive to future parameter displacement.
We derive a finite-displacement bound that relates task-specific
loss changes to FIM along the displacement
path.
For a prescribed displacement radius, reducing the worst-case
FIM trace over the surrounding neighborhood tightens the bound
on potential loss increase.
This result makes future retention an explicit consideration during
current-task learning and leads to our central design principle:
FIM should guide solution formation as well as
subsequent parameter protection.

We instantiate this principle with \textbf{Fisher-Shaped EWC
(FS-EWC)}, which shapes both how the current solution is learned
and how the retained reference is constructed.
As illustrated in Fig.~\ref{fig:fs_ewc_overview}, EWC constrains
subsequent parameter drift using historical importance, whereas
FS-EWC additionally shapes the solution before it is inherited by
future tasks.
During current-task learning, FS-EWC augments the EWC objective
with a neighborhood FIM-trace penalty, making sensitivity to
bounded future displacement an explicit optimization criterion
alongside task fitting and historical regularization.
At task completion, trajectory-based model merging constructs the
retained reference from multiple optimization states rather than
directly committing a single terminal iterate.
FIM is then estimated at the retained solution and
consolidated for subsequent protection.
The two components  address complementary stages of the
same process: shaping the solutions produced during learning and
constructing the solution passed forward for future.

Experiments across diverse vision and language benchmarks show that FS-EWC generally improves upon the corresponding
EWC baselines under both full-parameter and LoRA.
Controlled studies further show that shaping the solution itself
improves subsequent retention under the same protection rule,
demonstrating that future forgetting depends not only on how a solution
is protected but also on the solution being protected.
When combined with historical protection, 
solution shaping improves both retained-task and new-task accuracy in the controlled comparisons.
The gains also persist across alternative FIM computation and representations.
Together, these results suggest that the stability--plasticity
balance can be improved by jointly shaping solutions for future
retention and protecting acquired knowledge during subsequent learning.

Our contributions are summarized as follows:
\begin{itemize}
    \item
    We identify a limitation of EWC: it protects a learned solution
    using locally estimated parameter importance, but does not
    explicitly account for how sensitive that solution may become
    under finite future parameter drift.
    By analyzing task-loss changes along such displacement, we show
    that subsequent retention depends on Fisher sensitivity over the
    surrounding parameter neighborhood, motivating the solution itself
    to be shaped for future.

    \item
    We propose Fisher-Shaped EWC (FS-EWC), which extends EWC from
    protecting learned solutions to explicitly shaping and retaining
    solutions for future tasks.
    During task learning, FS-EWC reduces neighborhood FIM
    sensitivity while preserving historical EWC regularization.
    At task completion, trajectory-based model merging constructs the
    retained anchor, at which current-task FIM is
    estimated for subsequent protection.

% \item 
%     We articulate a broader principle for continual learning:
% effective knowledge retention requires considering not only how
% acquired knowledge is protected during subsequent learning, but also
% how current solutions are formed with future retention in mind.
% Experiments across vision and language benchmarks support this
% principle by showing consistent improvements over matched EWC
% baselines, improved subsequent retention under the same protection
% rule, and complementarity with alternative Fisher computation and representations.

\item
We propose a broader design principle for continual learning::
the stability--plasticity trade-off can be mitigated by incorporating
future retention into solution formation alongside historical
knowledge protection.
Experiments across vision and language benchmarks support this
principle, showing gains over corresponding EWC baselines and
complementarity with alternative FIM computation schemes and
matrix representations.

% argue that effective continual learning requires jointly
% considering how current solutions are formed for the future and
% how past knowledge is protected.
% Experiments across vision and language benchmarks support this view,
% showing improved performance over matched EWC baselines, improved
% retention under the same protection rule, and complementarity with
% alternative FIM computation and representations.

%     \item
%     We articulate a broader principle for continual learning:
% effective knowledge retention should be jointly incorporated into
% both the learning process and the subsequent preservation of acquired
% knowledge.
% Experiments across vision and language benchmarks support this
% principle by showing consistent improvements over matched EWC
% baselines, improved subsequent retention under the same protection
% rule, and complementarity with alternative Fisher computation
% schemes and matrix representations.
    % We articulate a broader principle for continual learning:
    % future knowledge retention should be explicitly incorporated into
    % the solution-learning process, rather than treated solely as a
    % post hoc protection problem after learning.
    % Experiments across vision and language benchmarks support this
    % principle by showing consistent improvements over matched EWC
    % baselines, improved subsequent retention under the same protection
    % rule, and complementarity with alternative Fisher computation
    % schemes and matrix representations.
\end{itemize}

\section{Related Works}

\textbf{Continual Learning.}
Continual learning seeks to preserve previously acquired knowledge
while enabling adaptation to evolving tasks and data distributions
\citep{Wang2024survey}.
% In class-incremental learning, new classes are introduced sequentially
% and all observed classes must be recognized without task identities at
% inference
% \citep{vandevenThreeTypesIncremental2022a,
% MasanaClassIncrementalLearningSurvey}.
A central challenge is the stability--plasticity trade-off:
restricting model changes can preserve previous knowledge but may
limit adaptation, whereas greater plasticity can accelerate forgetting.
Replay-based methods alleviate this tension by revisiting previous
examples, while replay-free learning must balance retention and
adaptation without access to historical data.
We focus on replay-free weight regularization, which addresses this
problem by constraining parameter updates during sequential learning.

\textbf{Elastic Weight Consolidation.}
Elastic Weight Consolidation (EWC) is a widely used regularization
baseline for replay-free continual learning \citep{EWC2017}.
% It retains a reference solution together with a Fisher-based estimate
% of parameter importance and constrains subsequent changes to parameters
% that are important for previously learned tasks.
Subsequent work has strengthened this protection mechanism by refining
how historical importance is accumulated and used across sequential
learning.
Online consolidation methods recursively update inherited importance
across tasks
\citep{huszar2018note,schwarz2018progress}, while trajectory-aware
approaches estimate parameter importance from the optimization process
\citep{SI2017Zenke,chaudhry2018riemannian}.
More recently, EWC-style regularization has been extended to low-rank
continual learning, demonstrating its applicability to LoRA-based
adaptation \citep{zheng2026ewclora}.
These approaches improve how acquired knowledge is protected during
subsequent learning, but largely treat the solution being protected as
given.
% Our work addresses the complementary question of how that solution
% should itself be learned and retained with future knowledge retention
% in mind.

% \paragraph{Elastic Weight Consolidation.}
% Elastic Weight Consolidation (EWC) is a widely used regularization
% baseline for replay-free continual learning \citep{EWC2017}.
% It retains a reference solution together with a Fisher-based estimate
% of parameter importance and constrains subsequent changes to parameters
% that are important for previously learned tasks.
% Subsequent work has strengthened this protection paradigm by refining
% how historical importance is accumulated and used across sequential
% learning.
% Online methods update the inherited
% importance summary across tasks
% \citep{huszar2018note,schwarz2018progress}, while trajectory-aware
% approaches estimate parameter importance from the optimization process
% % rather than only from the final solution
% \citep{SI2017Zenke,chaudhry2018riemannian}.
%  EWC regularization has also been extended to
% low-rank continual learning, demonstrating its applicability under
% LoRA-based adaptation \citep{zheng2026ewclora}.
% Despite their different formulations, these approaches primarily
% improve how a learned solution is subsequently protected.
% Our work addresses a complementary question: how the solution passed
% to future tasks should itself be learned and retained so that it is
% better prepared for subsequent knowledge preservation.

\textbf{FIM Estimation and Representation.}
The effectiveness of FIM-based  regularization method also depends on how
FIM is computed and represented.
Prior studies examine the distinction between empirical and
model-based FIM and the consequences of practical
computational approximations for continual learning
\citep{martens2020new,vandeven2025computation}.
Other work improves the representation of FIM beyond
diagonal approximations, including rotated and Kronecker-factored
structures
\citep{liu2018rotate,martens2015kfac}.
These advances improve the fidelity or expressiveness of the
importance signal used to protect a given reference solution.
Our work addresses an orthogonal dimension: how the reference solution
itself can be formed to better support future task.

\section{Methodology}
\label{sec:fisher_shaped_ewc}

% We extend EWC from protecting acquired solutions to considering how
% those solutions are learned and retained for future tasks.
% We first establish a finite-displacement sensitivity bound, then
% introduce neighborhood Fisher shaping and averaged anchors.

% The main text specifies the theoretical target and its computational
% surrogate; the numerical solver and complete learning procedure are
% provided in Appendix~\ref{app:fisher_numerical_details}.

\subsection{Preliminaries}
\label{sec:fs_problem_setup}

% \paragraph{Problem formulation and notation.}
\minisection{Problem formulation.} 
We consider replay-free class-incremental learning with tasks
$\sT_1,\ldots,\sT_T$. At task $t$, the learner receives
$\sS_t=\{(\vx_i^{(t)},y_i^{(t)})\}_{i=1}^{m_t}$ sampled from
$\mathcal D_t$. Inference distinguishes all observed classes without
task identities. Training accesses only $\sS_t$; historical examples
are neither retained nor replayed.
Let $f_{\vtheta}$ have trainable parameters $\vtheta\in\R^d$ and
predictive probabilities $p_{\vtheta}(y\mid\vx)$. Define
\begin{equation}
\begin{aligned}
\ell_{t,i}^{\mathrm{CE}}(\vtheta)
=-\log p_{\vtheta}(y_i^{(t)}\mid\vx_i^{(t)}), \qquad
\widehat{\mathcal L}_t^{\mathrm{CE}}(\vtheta)
=\frac1{m_t}\sum_{i=1}^{m_t}\ell_{t,i}^{\mathrm{CE}}(\vtheta).
\end{aligned}
\label{eq:task_empirical_ce_loss}
\end{equation}
The parameters retained after task $t$ are denoted by $\vtheta_t$.
% All gradients, perturbations, and importance matrices use the same
% trainable parameterization. 
% Appendix~\ref{app:notation} summarizes
% the notation.

% \paragraph{EWC and empirical Fisher.}
\minisection{EWC and empirical FIM.}
EWC \citep{EWC2017} retains a reference solution together with a FIM-based summary
of parameter importance, and penalizes subsequent changes along
directions that are important to previously learned tasks.
We adopt the online formulation
\citep{huszar2018note,schwarz2018progress}:
\begin{equation}
\begin{aligned}
\mathcal J_t^{\mathrm{EWC}}(\vtheta)
=
\widehat{\mathcal L}_t^{\mathrm{CE}}(\vtheta)
+\Omega_{t-1}(\vtheta),\quad
\Omega_{t-1}(\vtheta)
=
\frac{\lambda}{2}
(\vtheta-\vtheta_{t-1})^\top
\mLambda_{t-1}
(\vtheta-\vtheta_{t-1}).
\end{aligned}
\label{eq:online_ewc_objective}
\end{equation}
Here, $\vtheta_{t-1}$ is the retained reference from the preceding task,
and $\mLambda_{t-1}\succeq\mzero$ denotes the 
consolidated diagonal importance matrix.
% consolidated importance
% of previously learned parameters. 
Both are fixed during task-$t$
optimization, while $\lambda\geq0$ controls the strength of historical
protection.

For the current task, let
$\vg_{t,i}(\vtheta)=
\nabla_{\vtheta}\ell_{t,i}^{\mathrm{CE}}(\vtheta)$.
Define the empirical FIM as
\begin{equation}
\mF_t^{\mathrm{emp}}(\vtheta)
=
\frac{1}{m_t}
\sum_{i=1}^{m_t}
\vg_{t,i}(\vtheta)\vg_{t,i}(\vtheta)^\top.
\label{eq:empirical_fisher_definition}
\end{equation}
Its diagonal provides the task-specific importance contribution used
for subsequent EWC consolidation, while its trace will also serve below
as a scalar measure of local solution sensitivity. 
% Alternative Fisher
% computational approximations are discussed in
% Appendix~\ref{app:fisher_gradient_surrogate}.

% \paragraph{EWC and empirical Fisher.}
% EWC retains a reference solution together with a Fisher-based estimate
% of parameter importance and constrains subsequent changes to parameters
% that are important for previously learned tasks. We use its online formulation
% \citep{huszar2018note,onlineewc}:
% \begin{equation}
% \begin{aligned}
% \mathcal J_t^{\mathrm{EWC}}(\vtheta)
% =\widehat{\mathcal L}_t^{\mathrm{CE}}(\vtheta)
% +\Omega_{t-1}(\vtheta),\qquad
% \Omega_{t-1}(\vtheta)
% =\frac{\lambda}{2}(\vtheta-\vtheta_{t-1})^\top
% \mLambda_{t-1}(\vtheta-\vtheta_{t-1}).
% \end{aligned}
% \label{eq:online_ewc_objective}
% \end{equation}
% The inherited anchor $\vtheta_{t-1}$ and consolidated diagonal
% importance $\mLambda_{t-1}\succeq\mzero$ remain fixed during task-$t$
% training; $\lambda\geq0$ controls historical regularization.
% For $\vg_{t,i}=\nabla\ell_{t,i}^{\mathrm{CE}}$, the empirical Fisher is
% \begin{equation}
% \mF_t^{\mathrm{emp}}(\vtheta)
% =\frac1{m_t}\sum_{i=1}^{m_t}
% \vg_{t,i}(\vtheta)\vg_{t,i}(\vtheta)^\top.
% \label{eq:empirical_fisher_definition}
% \end{equation}

% Its diagonal provides the basis for current-task importance, with the
% damped contribution $\widehat{\mF}_t$ defined in
% Section~\ref{sec:fs_consolidation}.
% We use $\Diag$ to set off-diagonal matrix entries to zero.

\subsection{Solution Sensitivity to Future Drift}
\label{sec:fs_analysis}
\label{sec:finite_displacement_sensitivity}

The importance stored at an anchor does not by itself determine
its tolerance to finite future drift. For a candidate task-$t$
solution $\vtheta$, define the loss change induced by displacement
$\vDelta$ as
\begin{equation}
D_t(\vtheta,\vDelta)
=\widehat{\mathcal L}_t^{\mathrm{CE}}(\vtheta+\vDelta)
-\widehat{\mathcal L}_t^{\mathrm{CE}}(\vtheta).
\label{eq:future_loss_increase}
\end{equation}
Positive values describe deterioration in task-$t$ empirical loss.
For a later retained model $\vtheta_k$, $k>t$, the corresponding
displacement is $\vtheta_k-\vtheta_t$.
Since this displacement is unknown during task-$t$ learning, we
consider the prescribed neighborhood
$\sU_r(\vtheta)=\{\vtheta+\vepsilon:\|\vepsilon\|_2\leq r\}$.

\begin{proposition}[Finite-displacement sensitivity]
\label{prop:finite_displacement_fisher}
Suppose each $\ell_{t,i}^{\mathrm{CE}}$ is continuously differentiable
on an open set containing $\sU_r(\vtheta)$. Then every
$\|\vDelta\|_2\leq r$ satisfies
\begin{equation}
|D_t(\vtheta,\vDelta)|^2
\leq\int_0^1\vDelta^\top
\mF_t^{\mathrm{emp}}(\vtheta+\tau\vDelta)
\vDelta\,\mathrm d\tau.
\label{eq:integrated_fisher_bound}
\end{equation}
\end{proposition}
The proof is in Appendix~\ref{app:finite_displacement_fisher}.
% The bound does not require a stationary reference or identify the
% empirical Fisher with the loss Hessian.

To obtain a scalar criterion without specifying a displacement
direction, observe that
\begin{equation}
\vDelta^\top\mF_t^{\mathrm{emp}}(\vu)\vDelta
\leq\|\vDelta\|_2^2\frac1{m_t}
\sum_{i=1}^{m_t}\|\vg_{t,i}(\vu)\|_2^2.
\label{eq:fisher_directional_norm_bound}
\end{equation}
Accordingly, define the point FIM sensitivity
\begin{equation}
\mathcal R_{\mathrm F,t}(\vtheta)
=\frac1{m_t}\sum_{i=1}^{m_t}\|\vg_{t,i}(\vtheta)\|_2^2
=\tr\mF_t^{\mathrm{emp}}(\vtheta).
\label{eq:fisher_sensitivity_objective}
\end{equation}
Combining these expressions yields
\begin{equation}
|D_t(\vtheta,\vDelta)|^2
\leq\|\vDelta\|_2^2\int_0^1
\mathcal R_{\mathrm F,t}(\vtheta+\tau\vDelta)\,\mathrm d\tau.
\label{eq:integrated_trace_sensitivity_bound}
\end{equation}
This bound shows that finite-displacement loss change depends on
FIM sensitivity along the displacement path, rather than only at
the anchor.
Since the future path is unknown during task-$t$ learning, this
motivates controlling sensitivity over a bounded neighborhood around
the current solution, rather than relying only on importance estimated
at the retained anchor.
% This conservative bound depends on sensitivity throughout the
% segment, not only at the anchor. It motivates considering future
% retention when learning the current solution, rather than only
% estimating its importance after it has been learned.

\subsection{Fisher-Shaped EWC}
\label{sec:fs_framework}

FS-EWC combines neighborhood sensitivity shaping with an explicit rule for constructing the retained anchor. The former guides current-task learning while preserving the inherited EWC penalty; the latter selects the parameters at which current-task importance is estimated for subsequent learning.
A complementary distributional interpretation is provided in
Appendix~\ref{app:bayesian_ewc_surrogates}, while the complete FS-EWC algorithm
are detailed in Appendix~\ref{app:fisher_numerical_details}.

\subsubsection{Neighborhood Fisher Sensitivity Shaping}
\label{sec:neighborhood_fisher_objective}

% \subsubsection{Neighborhood Fisher Sensitivity Shaping}
% \label{sec:neighborhood_fisher_objective}

Over the prescribed neighborhood of radius $r$, we define the
worst-case FIM sensitivity as
\begin{equation}
\mathcal R_{\mathrm F,t}^{\max}(\vtheta;r)
=
\max_{\|\vepsilon\|_2\leq r}
\mathcal R_{\mathrm F,t}(\vtheta+\vepsilon)
=
\max_{\|\vepsilon\|_2\leq r}
\tr\mF_t^{\mathrm{emp}}(\vtheta+\vepsilon).
\label{eq:maximum_neighborhood_fisher}
\end{equation}
Here, $\vepsilon$ specifies locations at which sensitivity is evaluated,
whereas $\vDelta$ denotes possible future parameter drift.
The maximization characterizes the largest FIM sensitivity
within this neighborhood, without assuming a particular direction or
distribution of future drift.
% The maximization measures worst-case sensitivity over all
% perturbation directions within the prescribed radius $r$, without
% assuming a particular direction or distribution of future updates.

Combining Eq.~\eqref{eq:integrated_trace_sensitivity_bound} with
Young's inequality gives
\begin{equation}
\sup_{\|\vDelta\|_2\leq r}
|D_t(\vtheta,\vDelta)|
\leq
r\sqrt{\mathcal R_{\mathrm F,t}^{\max}(\vtheta;r)}
\leq
\mathcal R_{\mathrm F,t}^{\max}(\vtheta;r)
+\frac{r^2}{4}.
\label{eq:worst_case_fisher_loss_bound}
\end{equation}
Since the final term is independent of $\vtheta$, this motivates
\begin{equation}
\mathcal J_t^{\mathrm{FS\mbox{-}EWC}}(\vtheta;r)
=
\underbrace{
\widehat{\mathcal L}_t^{\mathrm{CE}}(\vtheta)
}_{\text{current-task fitting}}
+
\underbrace{
\mathcal R_{\mathrm F,t}^{\max}(\vtheta;r)
}_{\text{future-drift sensitivity shaping}}
+
\underbrace{
\Omega_{t-1}(\vtheta)
}_{\text{past-task protection}}.
\label{eq:fisher_shaping_objective}
\end{equation}

Let
$\vepsilon^\star\in
\arg\max_{\|\vepsilon\|_2\leq r}
\mathcal R_{\mathrm F,t}(\vtheta+\vepsilon)$.
At differentiable points, the gradient of
Eq.~\ref{eq:fisher_shaping_objective} takes the form
\begin{equation}
\nabla_{\vtheta}
\mathcal J_t^{\mathrm{FS\mbox{-}EWC}}
=
\vg_t(\vtheta)
+
\vh_t(\vtheta+\vepsilon^\star)
+
\lambda\mLambda_{t-1}
(\vtheta-\vtheta_{t-1}),
\label{eq:fs_ewc_gradient}
\end{equation}
% where
% $\vh_t(\vu)=\nabla_{\vu}\mathcal R_{\mathrm F,t}(\vu)$
% is the gradient of current-task Fisher sensitivity.
where
\begin{equation}
\vh_t(\vu)
=
\nabla_{\vu}\mathcal R_{\mathrm F,t}(\vu)
=
\frac{2}{m_t}
\sum_{i=1}^{m_t}
\mH_{t,i}(\vu)\vg_{t,i}(\vu).
\end{equation}
Thus, Fisher shaping introduces a Hessian--vector product correction
that reduces neighborhood sensitivity of the current-task solution,
while EWC retains historical parameter protection.

\subsubsection{Trajectory-Based Anchor Construction}
\label{sec:fs_consolidation}
Sensitivity shaping determines how the current-task solution is
learned, but continual learning also requires specifying which state
is passed to subsequent tasks.
Rather than retaining only the terminal iterate, we construct the
task-$t$ anchor by averaging selected learned checkpoints from its
optimization trajectory.
Let $\mathcal K_t$ denote the selected checkpoint indices, with
$|\mathcal K_t|=K$. We define
\begin{equation}
\overline{\vtheta}_t
=\frac1K\sum_{k\in\sK_t}\vtheta_{t,k},
\qquad\vtheta_t:=\overline{\vtheta}_t.
\label{eq:averaged_task_anchor}
\end{equation}
% All checkpoints are drawn from the same training trajectory and
% parameterization.
% The average can be accumulated online, the construction does not
% require storing the full optimization trajectory.
% To align the stored importance with the state actually inherited by
% future tasks, we estimate the current-task FIM at the
% retained anchor:
% % \begin{equation}
% $
% \widehat{\mF}_t
% =
% \Diag\!\left(\mF_t^{\mathrm{emp}}(\vtheta_t)\right)
% +\beta\mI,
% \quad \beta>0,
% \label{eq:fisher_at_averaged_solution}
% $
% % \end{equation}
% where $\beta$
% provides a small isotropic damping floor. 
% Following onlineEWC\citep{schwarz2018progress}, we normalize the FIM for each task and update
% $
% \mLambda_t
% =
% \mLambda_{t-1}
% +
% \widehat{\mF}_t.
% $
% % where $\gamma\in[0,1]$ controls the decay of historical importance.
% % \end{equation}
% % $.
% % where $\gamma$ is the importance-decay coefficient.
% % Only the current-task contribution is estimated at the new anchor;
% % historical data are not revisited.
% The pair $(\vtheta_t,\mLambda_t)$ initializes and regularizes the next
% task. 
The average can be accumulated online and does not require
storing the full optimization trajectory.
To align the stored importance with the state actually inherited by
future tasks, we estimate the current-task FIM at the retained anchor.
Following \citet{schwarz2018progress}, we normalize the current-task FIM
before consolidating it, and
additionally apply a small isotropic damping term before consolidation:
% \begin{equation}？
$
\mLambda_t
=
\mLambda_{t-1}
+
\widehat{\mF}_t,
\label{eq:averaged_anchor_consolidation}
% \end{equation}
$
where $\widehat{\mF}_t$ denotes the normalized and damped current-task
FIM.
The pair $(\vtheta_t,\mLambda_t)$ initializes and regularizes the next
task.

\section{Experiments}
\label{sec:experiments}

\subsection{Experimental Setup}
\label{sec:exp_setup}

\minisection{Models and datasets}
We evaluate our method on vision and language continual learning benchmarks.
Vision experiments use ViT-B/16~\citep{dosovitskiy2020image}
pretrained on ImageNet-21K.
We evaluate on CIFAR-100~\citep{krizhevsky2009learning},
ImageNet-R~\citep{ImagenetR_Hendrycks_2021_ICCV},
ImageNet-A~\citep{imagenet-a}, and DomainNet~\citep{Peng_2019_ICCV},
with ImageNet-A as the main benchmark.
Additional evaluations cover five fine-grained datasets:
FGVC Aircraft~\citep{maji2013finegrained},
Stanford Cars~\citep{krause2013cars},
CUB-200-2011~\citep{wah2011cub},
Oxford Flowers~\citep{nilsback2008automated},
and Oxford-IIIT Pets~\citep{parkhi2012cats},
along with a \textit{5Datasets} stream constructed from these datasets.
Language experiments use Llama-3.2-1B and Llama-3.2-3B~\citep{meta2024llama32}
on a continual text-classification benchmark comprising
AG News, Amazon Reviews, DBpedia, and
Yahoo Answers~\citep{NIPS2015_250cf8b5}. Detailed training configurations are
provided in Appendix~\ref{app:experimental_configurations}.

{%
\begin{figure*}[t]
\centering

% =========================================================
% Table
% =========================================================
\begin{minipage}{\textwidth}
\centering

\captionsetup{
    type=table,
    font=footnotesize,
    labelfont=bf,
    skip=1pt
}
\captionof{table}{
\textbf{Comparison results on ImageNet-R, ImageNet-A, DomainNet, and CIFAR-100.}
Results are reported in percentage with standard deviations.
Bold and underline indicate the highest and second-highest scores.
}
\label{tab:main_results_selected_benchmarks}

\scriptsize
\setlength{\tabcolsep}{2.6pt}
\renewcommand{\arraystretch}{0.96}

\begin{tabular}{@{}lcccccccc@{}}
\toprule
\multirow{2}{*}{Method}
& \multicolumn{2}{c}{ImageNet-R (T=10)}
& \multicolumn{2}{c}{ImageNet-A (T=10)}
& \multicolumn{2}{c}{DomainNet (T=5)}
& \multicolumn{2}{c}{CIFAR-100 (T=10)} \\
\cmidrule(lr){2-3}
\cmidrule(lr){4-5}
\cmidrule(lr){6-7}
\cmidrule(lr){8-9}
& FAA & AAA
& FAA & AAA
& FAA & AAA
& FAA & AAA \\
\midrule

Joint Train
& \ExpRes{81.69}{0.28} & \ExpRes{86.25}{0.09}
& \ExpRes{65.01}{0.74} & \ExpRes{74.10}{0.44}
& \ExpRes{76.84}{0.06} & \ExpRes{81.25}{0.05}
& \ExpRes{92.82}{0.20} & \ExpRes{95.41}{0.05} \\

Finetune
& \ExpRes{60.42}{1.64} & \ExpRes{73.18}{0.32}
& \ExpRes{16.31}{7.89} & \ExpRes{30.44}{13.18}
& \ExpRes{51.46}{0.47} & \ExpRes{67.08}{1.13}
& \ExpRes{69.49}{0.50} & \ExpRes{80.35}{0.87} \\

L2P
& \ExpRes{71.26}{0.44} & \ExpRes{76.13}{0.46}
& \ExpRes{42.94}{1.27} & \ExpRes{51.40}{1.95}
& \ExpRes{70.26}{0.25} & \ExpRes{75.83}{0.98}
& \ExpRes{83.18}{1.20} & \ExpRes{87.69}{1.05} \\

DualPrompt
& \ExpRes{68.22}{0.20} & \ExpRes{73.81}{0.39}
& \ExpRes{45.49}{0.96} & \ExpRes{54.68}{1.24}
& \ExpRes{68.26}{0.90} & \ExpRes{73.84}{0.45}
& \ExpRes{81.48}{0.86} & \ExpRes{86.41}{0.66} \\

EWC
& \ExpRes{64.16}{0.92} & \ExpRes{75.27}{0.55}
& \ExpRes{21.30}{4.79} & \ExpRes{45.77}{0.43}
& \ExpRes{58.85}{2.92} & \ExpRes{72.12}{0.34}
& \ExpRes{75.06}{3.49} & \ExpRes{85.72}{1.04} \\

FS-EWC
& \ExpRes{68.46}{1.37} & \ExpRes{76.60}{2.45}
& \ExpRes{26.89}{0.98} & \ExpRes{46.08}{2.37}
& \ExpRes{64.77}{0.53} & \ExpRes{74.87}{0.59}
& \ExpRes{77.32}{0.59} & \ExpRes{85.70}{0.27} \\

\midrule

Vanilla LoRA
& \ExpRes{64.87}{0.73} & \ExpRes{75.57}{0.23}
& \ExpRes{40.01}{1.32} & \ExpRes{58.28}{0.85}
& \ExpRes{69.79}{0.11} & \ExpRes{77.44}{0.08}
& \ExpRes{82.99}{0.84} & \ExpRes{89.74}{0.58} \\

BiLoRA
& \ExpRes{74.28}{0.92} & \ExpRes{77.38}{0.88}
& \ExpRes{51.05}{0.74} & \ExpRes{62.82}{0.65}
& \ExpRes{69.75}{0.23} & \ExpRes{73.86}{0.21}
& \ExpRes{85.99}{0.49} & \ExpRes{90.62}{0.42} \\

% CL-LoRA
% & \ExpRes{\mathbf{78.72}}{0.44}
% & \ExpRes{\mathbf{85.20}}{0.45}
% & \ExpRes{\underline{57.62}}{0.89}
% & \ExpRes{\underline{70.76}}{0.63}
% & \ExpRes{71.06}{0.31} & \ExpRes{77.76}{0.29}
% & \ExpRes{87.65}{0.53} & \ExpRes{\underline{92.25}}{0.41}
%  \\

InfLoRA
& \ExpRes{\underline{74.41}}{0.63} & \ExpRes{\underline{80.31}}{0.60}
& \ExpRes{50.75}{1.33} & \ExpRes{64.36}{1.01}
& \ExpRes{71.01}{0.05} & \ExpRes{77.75}{0.03}
& \ExpRes{86.34}{0.76} & \ExpRes{91.33}{0.48} \\

SD-LoRA
& \ExpRes{72.93}{2.76} & \ExpRes{79.80}{0.15}
& \ExpRes{55.23}{0.94} & \ExpRes{66.10}{0.54}
& \ExpRes{71.27}{0.14} & \ExpRes{77.70}{0.07}
& \ExpRes{86.77}{0.30} & \ExpRes{90.96}{0.30} \\

EWCLoRA
& \ExpRes{72.58}{0.58} & \ExpRes{78.38}{0.69}
& \ExpRes{\underline{56.42}}{0.88} & \ExpRes{\underline{67.09}}{0.62}
& \ExpRes{\underline{73.38}}{0.14}
& \ExpRes{\underline{79.50}}{0.13}
& \ExpRes{\underline{87.91}}{0.11}
& \ExpRes{\textbf{92.27}}{0.19} \\

FS-EWCLoRA
& \ExpRes{\textbf{79.81}}{0.21}
& \ExpRes{\textbf{83.51}}{0.11}
& \ExpRes{\mathbf{62.08}}{1.62}
& \ExpRes{\textbf{71.66}}{0.25}
& \ExpRes{\mathbf{73.95}}{0.04}
& \ExpRes{\mathbf{79.50}}{0.04}
& \ExpRes{\mathbf{88.03}}{0.10}
& \ExpRes{\underline{92.00}}{0.16} \\

\bottomrule
\end{tabular}
\end{minipage}

% =========================================================
% Exact spacing between table and figure
% =========================================================
\vspace{+4pt}

% =========================================================
% Figure
% =========================================================
\begin{minipage}{\textwidth}
\centering

\includegraphics[
    width=\textwidth,
    trim=0 0 0 0,
    clip
]{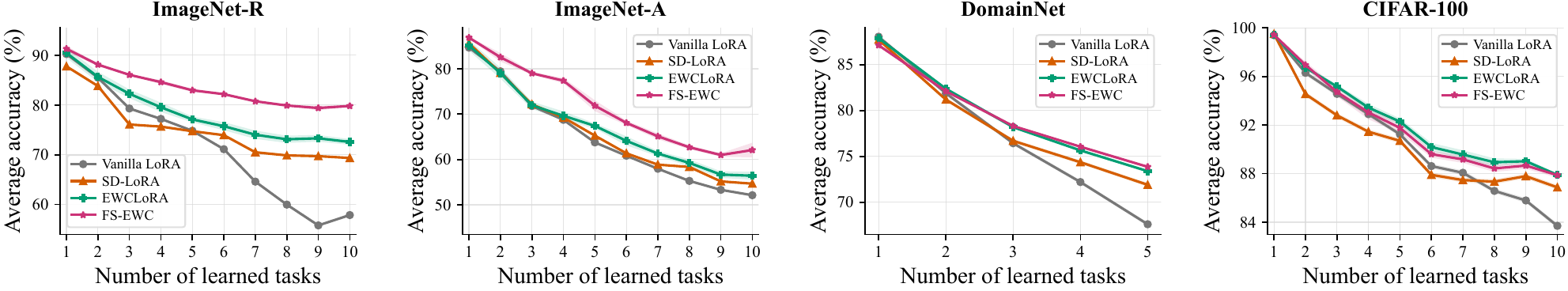}

% \vspace{-6pt}

\captionsetup{
    type=figure,
    font=footnotesize,
    labelfont=bf,
    skip=1pt
}
\captionof{figure}{
% Task-wise performance comparison of different methods across various datasets.
% \textbf{Continual-learning performance across learning stages.}
Stage-wise average accuracy $A_t$ over all learned tasks 
under LoRA on four vision benchmarks.}
\label{fig:taskwise_performance}

\end{minipage}

\end{figure*}
}

\minisection{Metrics}
We report stage-wise average accuracy $A_t$ over all tasks observed
through learning stage $t$, together with final average accuracy
(FAA), stage-averaged accuracy (AAA), and backward transfer (BWT).
% FAA summarizes performance after the complete task sequence, AAA
% summarizes performance throughout sequential learning, and BWT
% measures changes in previously learned tasks.
For the solution-shaping diagnostics, we additionally measure
task-specific loss and accuracy, effective-weight displacement,
empirical Fisher sensitivity, and displacement tolerance.
We distinguish realized endpoint displacement $d_t$ from controlled
diagnostic displacement $d$, with
$D_{1,t}^{\mathrm{test}}$ and $\overline D_1^{+}(d)$ denoting the
corresponding test-loss change and mean positive training-loss
increase.
Formal definitions and measurement protocols are provided in
Appendix~\ref{app:diagnostic_metrics}.
\subsection{Main result}
\minisection{Overall Performance across Benchmarks and Model Configurations.}
We first evaluate whether the benefit of FS-EWC persists across
training parameterizations, task domains, and model scales.

Table~\ref{tab:main_results_selected_benchmarks} reports results on
the four standard vision benchmarks, with the corresponding
LoRA-based performance trajectories shown in
Fig.~\ref{fig:taskwise_performance}.
FS-EWC improves FAA over the corresponding EWC baseline across
all four benchmarks under both full-parameter fine-tuning and
LoRA.
The LoRA results on ImageNet-R and ImageNet-A are particularly
notable.
On ImageNet-R, EWCLoRA trails several competing LoRA methods,
whereas FS-EWCLoRA improves FAA
to $79.81\%$, surpassing all compared LoRA baselines.
On ImageNet-A, EWCLoRA already leads the competing LoRA baselines,
yet FS-EWCLoRA achieves a further gain,
raising FAA from $56.42\%$ to $62.08\%$.
These results suggest that EWC parameter protection alone
can be insufficient in challenging settings, whereas jointly
shaping and protecting task solutions can yield substantial
performance gains.

We next examine more heterogeneous visual recognition settings in Table~\ref{tab:exp_fine}.
Across Aircraft, Cars, CUB-200, Flowers, and 5Datasets,
FS-EWCLoRA improves FAA over its matched EWCLoRA baseline in every
case.
It achieves the highest FAA on Aircraft, Cars, and 5Datasets, and remains stronger on CUB-200 and Flowers.
Thus, the improvement over EWC persists across substantially different
class structures and task compositions.

{%
\setlength{\textfloatsep}{6pt}
\setlength{\abovecaptionskip}{2pt}
\setlength{\belowcaptionskip}{2pt}

\begin{table*}[t]
\centering
\captionsetup{type=table,font=footnotesize,labelfont=bf}
% =========================================================
% Table 1: Vision
% =========================================================
\caption{
\textbf{Fine-grained and heterogeneous vision performance}.
FAA and stage-averaged AAA are reported in percentages.
5Datasets denotes the cross-dataset continual learning stream in the
following order:
CUB-200 $\rightarrow$ Aircraft $\rightarrow$ Cars $\rightarrow$
Flowers $\rightarrow$ Pets.
The corresponding average-accuracy trajectories are
provided in Fig.~\ref{fig:finegrained_stagewise}.
}
\label{tab:exp_fine}

\scriptsize
\setlength{\tabcolsep}{2.2pt}
\renewcommand{\arraystretch}{1.08}

\begin{tabular*}{\linewidth}{@{\extracolsep{\fill}}l*{10}{r}@{}}
\toprule
Method
& \multicolumn{2}{c}{FGVC-Aircraft}
& \multicolumn{2}{c}{Stanford Cars}
& \multicolumn{2}{c}{CUB-200}
& \multicolumn{2}{c}{Oxford Flowers}
& \multicolumn{2}{c}{5Datasets}\\
\cmidrule(lr){2-3}
\cmidrule(lr){4-5}
\cmidrule(lr){6-7}
\cmidrule(lr){8-9}
\cmidrule(lr){10-11}
& FAA & AAA
& FAA & AAA
& FAA & AAA
& FAA & AAA
& FAA & AAA\\
\midrule

Vanilla LoRA
% & 39.66 & 39.51
% & 28.52 & 39.41
% & 57.79 & 70.00
% & 53.05 & 66.90
% & 66.80 & 68.81\\
% % Vanilla LoRA & \ExpRes{39.85}{0.78} & \ExpRes{38.42}{1.19} & \ExpRes{31.74}{2.57} & \ExpRes{41.67}{2.07} & \ExpRes{58.31}{1.72} & \ExpRes{69.61}{0.57} & \ExpRes{52.92}{1.27} & \ExpRes{66.49}{1.29} & \ExpRes{66.90}{1.46} & \ExpRes{68.85}{0.24}\\
& 39.85 & 38.42 & 31.74 & 41.67 & 58.31 & 69.61 & 52.92 & 66.49 & 66.90 & 68.85\\

InfLoRA
% & 40.96 & 41.23
% & 34.34 & 42.20
% & 60.61 & 70.60
% & 60.06 & 72.98
% & \underline{77.75} & \underline{77.02}\\
% & \ExpRes{41.35}{0.29} & \ExpRes{39.62}{0.59} & \ExpRes{35.98}{2.28} & \ExpRes{44.57}{1.98} & \ExpRes{61.36}{0.42} & \ExpRes{70.38}{0.53} & \ExpRes{57.34}{3.38} & \ExpRes{69.98}{2.97} & \ExpRes{78.77}{1.08} & \ExpRes{77.25}{0.67}\\
& \underline{41.35} & \underline{39.62} &  \underline{35.98} &  \underline{44.57} & 61.36 & 70.38 & 57.34 & 69.98 & 78.77 & 77.25\\

OLoRA
% & 36.10 & 37.81
% & 32.85 & 42.30
% & 62.38 & 70.17
% & 48.90 & 64.26
% & 77.58 & 74.52\\
 % \ExpRes{37.67}{1.22} & \ExpRes{37.77}{0.88} & \ExpRes{35.30}{0.53} & \ExpRes{44.70}{0.67} & \ExpRes{60.70}{1.26} & \ExpRes{68.71}{0.52} & \ExpRes{47.68}{1.32} & \ExpRes{63.18}{1.17} & \ExpRes{79.12}{0.45} & \ExpRes{76.96}{0.13}\\
& 37.67 & 37.77 & 35.30 & 44.70 & 60.70 & 68.71 & 47.68 & 63.18 & 79.12 & 76.96\\

SDLoRA
& 25.96 & 26.63
& {34.89} &{43.28}
& 62.01 & 69.52
& 49.05 & 64.59
& 74.42 & 74.10\\

CL-LoRA
% & 29.28 & 26.53
% & 34.51 & 42.17
% & \textbf{78.29} & \textbf{80.88}
% & \textbf{97.82} & \textbf{98.33}
% & 66.98 & 61.99\\
% & \ExpRes{29.35}{0.06} & \ExpRes{26.54}{0.06} & \ExpRes{34.01}{0.63} & \ExpRes{41.86}{0.30} & \ExpRes{78.31}{0.07} & \ExpRes{81.01}{0.12} & \ExpRes{97.83}{0.005} & \ExpRes{98.34}{0.002} & \ExpRes{67.05}{0.07} & \ExpRes{61.97}{0.08}\\
& 29.35 & 26.54 & 34.01 & 41.86 & \textbf{78.31} & \textbf{81.01} & \textbf{97.83} & \textbf{98.34} & 67.05 & 61.97\\

EWCLoRA
% & \underline{41.83} & \underline{39.70}
% & 34.56 & 43.94
% & 60.41 & 69.40
% & 53.43 & 65.45
% & 73.87 & 73.07\\
%  & \ExpRes{40.64}{1.61} & \ExpRes{38.44}{1.10} & -- & -- & \ExpRes{61.34}{0.91} & \ExpRes{70.36}{1.38} & \ExpRes{52.02}{1.10} & \ExpRes{65.36}{0.32} & \ExpRes{73.30}{0.57} & \ExpRes{72.93}{0.24}\\
& 40.64 & 38.44 & 34.56 & 43.94 & 61.34 & 70.36 & 52.02 & 65.36 & 73.30 & 72.93\\

FS-EWCLoRA
% & \textbf{46.66} & \textbf{43.65}
% & \textbf{53.04} & \textbf{56.26}
% & \underline{74.22} & \underline{76.47}
% & \underline{86.75} & \underline{91.76}
% & \textbf{83.37} & \textbf{81.19}\\

% \ExpRes{45.97}{1.02}$^{\dagger}$ & \ExpRes{43.41}{0.38}$^{\dagger}$ & \ExpRes{53.04}{2.02} & \ExpRes{56.26}{1.52} & \ExpRes{73.42}{0.97} & \ExpRes{76.26}{0.38} & \ExpRes{84.68}{1.96} & \ExpRes{90.00}{0.17} & \ExpRes{83.24}{0.38} & \ExpRes{81.19}{0.28}\\

& \textbf{46.66} & \textbf{43.41} & \textbf{53.04} & \textbf{56.26} & \underline{73.42} & \underline{76.26} & \underline{84.68} & \underline{90.00} & \textbf{83.24} &\textbf{81.19}\\

\bottomrule
\end{tabular*}

% =========================================================
% Reduce the vertical gap between the two tables
% =========================================================
\vspace{+4pt}

% =========================================================
% Table 2: NLP
% =========================================================
\captionsetup{type=table,font=footnotesize,labelfont=bf}
\caption{
\textbf{Continual NLP benchmark performance of Llama-3.2-1B and
Llama-3.2-3B across three task orders}.
FAA (\%) is reported for each order, and Avg. denotes the average
FAA across the three orders.
Task-order are given in Table~\ref{tab:nlp_task_orders},
and the corresponding trajectories are
provided in Fig.~\ref{fig:llama_taskwise_combined_v6}.
}
\label{tab:exp_nlp}

\scriptsize
\setlength{\tabcolsep}{3.6pt}
\renewcommand{\arraystretch}{1.08}

\begin{tabular*}{\linewidth}{
    @{\extracolsep{\fill}}l*{8}{c}@{}
}
\toprule
\multirow{2}{*}{Method}
& \multicolumn{4}{c}{Llama-3.2-1B}
& \multicolumn{4}{c}{Llama-3.2-3B} \\
\cmidrule(lr){2-5}
\cmidrule(lr){6-9}
& Order-1 & Order-2 & Order-3 & Avg.
& Order-1 & Order-2 & Order-3 & Avg. \\
\midrule

SeqLoRA
% & 66.90 & 55.71 & 68.06 & 63.56
% & 67.86 & 67.26 & 67.40 & 67.51 \\

& \ExpRes{65.55}{0.59} & \ExpRes{58.53}{0.86}
& \ExpRes{67.83}{1.65} & \ExpRes{63.97}{0.96}
% & -- & -- & -- & -- \\
% & 67.21 (0.92) & 67.00 (0.37) & 69.59 (3.10) & 67.93 (0.60)
& \ExpRes{67.21}{0.92} & \ExpRes{67.00}{0.37} & \ExpRes{69.59}{3.10} & \ExpRes{67.93}{0.60} \\

IncLoRA
% & 64.52 & 56.19 & 65.44 & 62.05
% & 65.23 & 66.74 & 69.55 & 67.17 \\

& \ExpRes{66.20}{0.56} & \ExpRes{61.72}{2.47}
& \ExpRes{65.59}{2.46} & \ExpRes{64.50}{1.45}
& \ExpRes{66.96}{2.45} & \ExpRes{68.47}{2.45} & \ExpRes{70.58}{1.45}
& \ExpRes{68.67}{2.12} \\

OLoRA
% & 65.81 & 58.28 & 66.92 & 63.67
% & 71.62 & 70.94 & 72.05 & 71.54 \\
& \ExpRes{66.93}{1.13} & \ExpRes{61.52}{1.26}
& \ExpRes{\underline{67.28}}{2.17} & \ExpRes{65.24}{1.23}
& \ExpRes{\underline{71.19}}{1.24} & \ExpRes{\underline{71.12}}{0.11}
& \ExpRes{\underline{73.06}}{0.66} & \ExpRes{\underline{71.79}}{0.51} \\

TreeLoRA
% & 66.93 & 61.53 & 66.16 & 64.88
% & 72.02 & 69.01 & \textbf{73.46} & 71.50 \\
& \ExpRes{\underline{68.40}}{1.27} & \ExpRes{\underline{63.81}}{2.25}
& \ExpRes{66.60}{1.36} & \ExpRes{\underline{66.27}}{1.45}
& \ExpRes{70.56}{1.27} & \ExpRes{69.97}{0.83}
& \ExpRes{72.92}{0.53} & \ExpRes{71.15}{0.32} \\

EWCLoRA
% & 67.19 & 61.92 & 66.54 & 65.22
% & 68.36 & 69.18 & {73.04} & 70.19 \\
& \ExpRes{66.35}{0.84} & \ExpRes{62.14}{2.56}
& \ExpRes{65.62}{2.23} & \ExpRes{64.70}{1.48}
& \ExpRes{68.59}{0.24} & \ExpRes{70.28}{0.39}
& \ExpRes{72.07}{0.76} & \ExpRes{70.31}{0.31} \\

% FS-EWCLoRA
% & \textbf{69.85} & \textbf{68.13} & \textbf{70.62} & \textbf{69.53}
% & \textbf{72.71} & \textbf{73.51} & {72.66} & \textbf{72.96} \\
FS-EWCLoRA
& \ExpRes{\textbf{69.57}}{0.27} & \ExpRes{\textbf{67.54}}{1.04}
& \ExpRes{\textbf{70.35}}{0.25} & \ExpRes{\textbf{69.15}}{0.49}
& \ExpRes{\textbf{72.57}}{0.47} & \ExpRes{\textbf{72.73}}{0.67}
& \ExpRes{\textbf{73.93}}{0.23} & \ExpRes{\textbf{73.08}}{0.40} \\

\bottomrule
\end{tabular*}
\vspace{-2pt}
\end{table*}
}

Finally, we evaluate continual text classification with
Llama-3.2-1B and Llama-3.2-3B across three task orders in Table~\ref{tab:exp_nlp}.
FS-EWCLoRA increases the mean FAA over EWCLoRA from $64.70\%$ to
$69.15\%$ on Llama-3.2-1B and from $70.31\%$ to $73.08\%$ on
Llama-3.2-3B, with improvements in five of the six model--order
comparisons.
These results indicate that the benefit is not restricted to vision
or to a single model scale.

\minisection{How Does Solution Shaping Affect the
Stability--Plasticity Trade-off?}
\label{sec:exp_rq2}
% We next examine whether the aggregate gains above correspond to a
% more favorable stability--plasticity trade-off.
% In particular, we ask whether improved retention requires stronger
% restriction of future parameter movement, or whether shaping the
% retained solution can make such movement less damaging.
We next examine whether the aggregate gains above reflect a more
favorable stability--plasticity trade-off.
Specifically, we ask whether improved stability is achieved merely by
restricting future parameter movement, or by shaping the retained
solution so that subsequent adaptation can proceed with less forgetting.
To isolate the effect of Fisher shaping,  each task
only retains its terminal iterate without merging.

Fig.~\ref{fig:realized_future_drift} first examines retention under
actual subsequent learning.
FS-EWCLoRA does not achieve improved stability by restricting
parameter movement more strongly.
At the final stage, its effective-weight displacement is larger than
that of EWCLoRA ($1.442$ versus $1.254$), while the Task~1 test-loss
increase is substantially smaller ($0.033$ versus $0.229$) and
Task~1 accuracy is higher ($86.64\%$ versus $81.11\%$).
The corresponding path Fisher is also much
lower ($0.015$ versus $0.240$).
Thus, the retention gain is achieved despite greater realized
movement, indicating that improved stability does not require stronger
suppression of subsequent updates.
This behavior is consistent with the finite-displacement analysis,
which attributes old-task degradation jointly to displacement
magnitude and the sensitivity encountered along that displacement.

% Figure~\ref{fig:realized_future_drift} first tracks how the retained
% Task~1 solution changes during actual subsequent learning.
% FS-EWCLoRA does not achieve better retention by restricting parameter
% movement more strongly.
% At the final stage, its effective-weight displacement is larger than
% that of EWCLoRA ($1.442$ versus $1.254$), while the Task~1 test-loss
% increase is substantially smaller ($0.033$ versus $0.229$) and
% Task~1 accuracy is higher ($86.64\%$ versus $81.11\%$).
% The corresponding directional path Fisher sensitivity is also much
% lower ($0.015$ versus $0.240$).
% Thus, the retention difference cannot be explained by displacement
% magnitude alone; lower degradation can coexist with larger movement
% when the retained solution remains less sensitive along the resulting
% displacement directions. This behavior is consistent with the finite-displacement analysis,
% which separates the magnitude of future drift from the Fisher
% sensitivity encountered along that displacement.

% \begin{figure}[h]
%     \centering
%     \includegraphics[width=1\linewidth]{Figure_A_realized_future.pdf}
%      \caption{
%     Realized drift diagnostics on ImageNet-A.
%     Starting from the retained Task~1 solution, we track the weight displacement $d_t$, Task~1 loss change
%     $D_{1,t}^{\mathrm{test}}$, Task~1 test accuracy, and the empirical-Fisher 
%     % $\overline{\mathcal S}_{\mathrm F,1}^{\mathrm{path}}(t)$ 
%     along the line connecting the Task~1 solution to each subsequent
%     task endpoint.}
%     \label{fig:realized_future_drift}
% \end{figure}

{%
\setlength{\abovecaptionskip}{2pt}
\setlength{\belowcaptionskip}{0pt}

\begin{figure}[t]
\centering

\includegraphics[width=\linewidth]
{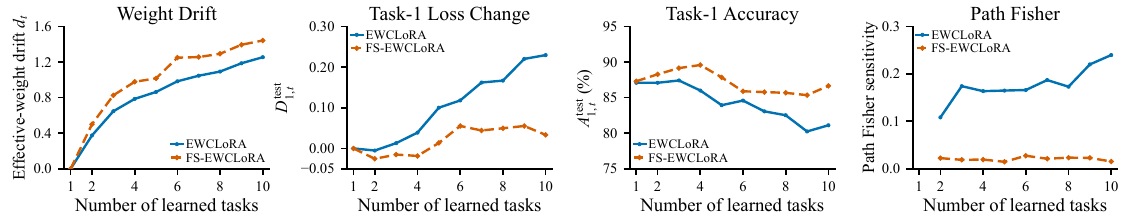}
\captionsetup{font=footnotesize,labelfont=bf}
% \caption{
% Realized drift diagnostics on ImageNet-A.
% Starting from the retained Task~1 solution, we track the weight
% displacement $d_t$, Task~1 loss change $D_{1,t}^{\mathrm{test}}$,
% Task~1 test accuracy, and the empirical-Fisher along the line
% connecting the Task~1 solution to each subsequent task endpoint.
% }
\caption{
\textbf{Less forgetting despite larger realized drift.}
Realized-drift diagnostics on ImageNet-A with LoRA.
Relative to the Task 1 solution, we report effective-weight
displacement $d_t$, Task 1 loss change
$D_{1,t}^{\mathrm{test}}$, Task 1 accuracy, and empirical-Fisher sensitivity
averaged along the straight segment to each subsequent task endpoint.
% FS-EWCLoRA better retains Task 1 despite larger endpoint displacement,
% consistent with lower sensitivity along these diagnostic segments.
}

\label{fig:realized_future_drift}

\vspace{+3pt}

\includegraphics[width=\linewidth]
{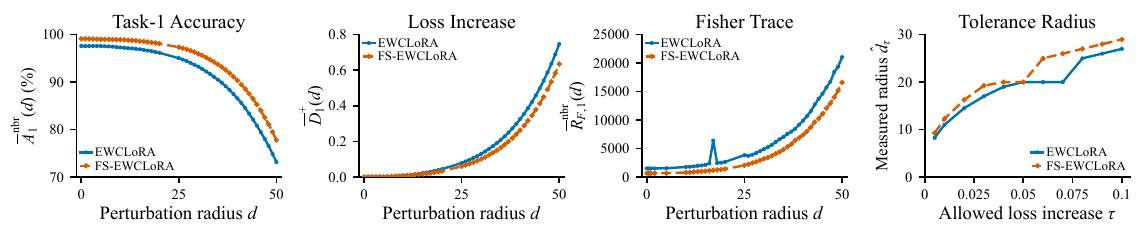}
\captionsetup{font=footnotesize,labelfont=bf}
% \caption{
% Controlled-neighborhood diagnostics around the retained Task~1
% solution on ImageNet-A.
% Shared random perturbations of radius $d$ are applied in the effective
% LoRA weight.
% We report Task~1 training accuracy, loss increase,
% averaged FIM trace, and the measured tolerance radius
% under loss budget $\tau$.
% }
\caption{
\textbf{Lower neighborhood sensitivity and greater empirical
displacement tolerance.}
Controlled-neighborhood diagnostics around the retained Task 1
solutions on ImageNet-A.
Shared random directions are used to apply perturbations of radius
$d$ in the weight space.
We report Task 1 training accuracy, mean positive
training-loss increase, mean empirical-Fisher trace, and the measured
tolerance radius under loss budget $\tau$.
% FS-EWCLoRA exhibits lower loss increase and Fisher
% sensitivity, and admits larger radii under the
% same budget.
% Tolerance is evaluated over the sampled directions and radius grid,
% rather than certified over the continuous neighborhood.
}
\label{fig:controlled_neighborhood}

\end{figure}
}

To isolate the role of solution sensitivity, we next control the
displacement magnitude directly.
Fig.~\ref{fig:controlled_neighborhood} perturbs the retained
Task~1 solutions along matched random directions in the same
effective LoRA weight.
For the same perturbation radius, FS-EWCLoRA exhibits a smaller mean
positive loss increase together with a lower empirical-Fisher trace
throughout the measured range.
Accordingly, for a fixed empirical retention budget, the shaped
solution admits a larger measured displacement radius: for example,
$\widehat d_\tau$ increases from $11$ to $12.25$ at
$\tau=0.01$, and is larger for ten of the eleven evaluated budgets.
Hence, shaping enlarges the empirically retention-compatible region
around the learned solution, providing more room for subsequent
adaptation under the same retention tolerance.

Table~\ref{tab:two_task_lora_diagnostics_imageneta_raw20000}
further separates solution shaping from historical protection.
Comparing Vanilla LoRA with EWCLoRA shows the characteristic effect
of EWC: the effective displacement $d_2$ decreases from $0.38$ to
$0.28$, accompanied by a modest improvement in retained Task~1
accuracy , while Task~2 accuracy decreases
slightly.
Thus, historical protection primarily improves stability by
restricting subsequent parameter movement, with a small cost to
current-task adaptation.
% Fisher shaping changes this behavior.

% Together with Figs.~\ref{fig:realized_future_drift}
% and~\ref{fig:controlled_neighborhood}, this shows that
However, Fisher shaping improves
the retention--adaptation balance without relying on smaller parameter
movement.
Relative to Vanilla LoRA, Fisher Shaping improves both retained-task
and current-task accuracy despite increasing $d_2$ from
$0.38$ to $0.50$.
The same pattern holds with EWC protection:
FS-EWCLoRA improves $a_{21}$ from $82.95\%$ to $83.50\%$ and
$a_{22}$ from $76.25\%$ to $79.92\%$, while $d_2$ increases from
$0.28$ to $0.43$.
The stage-specific variants clarify why.
Shaping Task~1 reduces the endpoint sensitivity proxy
$tr \mF_1^{\mathrm{emp}}(\vtheta)$ from $31.44$ to $21.93$ and substantially improves its
subsequent retention under either unprotected training
($82.41\%\!\rightarrow\!86.54\%$) or the same
$\mathsf C+\mathsf P$ protection
($82.95\%\!\rightarrow\!86.32\%$).
Thus, shaping improves the solution passed forward for future
retention.
Conversely, shaping during Task~2 substantially improves current-task
adaptation ($76.25\%\!\rightarrow\!81.89\%$), although this shifts
the balance away from immediate Task~1 retention.
These results indicate that Fisher shaping serves two related roles:
it forms solutions that are more suitable for subsequent retention
and shapes the current-task solution during adaptation.
Combined with historical protection, these effects yield a more
favorable stability--plasticity balance than EWC alone.

\begin{table*}[t]
\centering
\captionsetup{font=footnotesize,labelfont=bf}
\caption{
\textbf{Two-task diagnostics of solution shaping and historical protection
with LoRA on ImageNet-A.}
$\mathsf C$, $\mathsf F$, and $\mathsf P$ denote current-task fitting, Fisher shaping, and historical EWC protection,
respectively.
}
\label{tab:two_task_lora_diagnostics_imageneta_raw20000}

\scriptsize
\setlength{\tabcolsep}{1.8pt}
\renewcommand{\arraystretch}{1.06}

\begin{tabular}{
@{}l
@{\hspace{3pt}}cc
@{\hspace{3pt}}cc
@{\hspace{3pt}}cccc
@{\hspace{3pt}}c
@{}
}
\toprule
& \multicolumn{2}{c}{Training objective}
& \multicolumn{2}{c}{Task 1 endpoint}
& \multicolumn{4}{c}{Task 2 endpoint}
& \shortstack{Task 2update} \\
\cmidrule(lr){2-3}
\cmidrule(lr){4-5}
\cmidrule(lr){6-9}

Variant
& Task 1
& Task 2
& $a_{11}\uparrow$
& $\tr \mF_1^{\mathrm{emp}}$
& $a_{21}\uparrow$
& $a_{22}\uparrow$
& BWT$\uparrow$
& $A_2\uparrow$
& $d_2$ \\
\midrule

Vanilla LoRA
& $\mathsf C$
& $\mathsf C$
& \ExpRes{87.08}{0.38}
& \ExpRes{31.44}{4.24}
& \ExpRes{82.41}{1.13}
& \ExpRes{76.38}{0.40}
& \ExpRes{-4.67}{1.36}
& \ExpRes{79.39}{0.74}
& \ExpRes{0.38}{0.02} \\

Fisher Shaping
& $\mathsf C+\mathsf F$
& $\mathsf C+\mathsf F$
& \ExpRes{87.30}{0.57}
& \ExpRes{21.93}{2.98}
& \ExpRes{83.39}{0.33}
& \ExpRes{80.84}{1.14}
& \ExpRes{-3.91}{0.65}
& \ExpRes{82.12}{0.59}
& \ExpRes{0.50}{0.01} \\

Task-1 only
& $\mathsf C+\mathsf F$
& $\mathsf C$
& \ExpRes{87.30}{0.57}
& \ExpRes{21.93}{2.98}
& \ExpRes{86.54}{0.82}
& \ExpRes{72.96}{1.38}
& \ExpRes{-0.76}{1.36}
& \ExpRes{79.75}{1.10}
& \ExpRes{0.36}{0.01} \\

\midrule

EWCLoRA
& $\mathsf C$
& $\mathsf C+\mathsf P$
& \ExpRes{87.08}{0.38}
& \ExpRes{31.44}{4.24}
& \ExpRes{82.95}{0.76}
& \ExpRes{76.25}{0.60}
& \ExpRes{-4.13}{1.00}
& \ExpRes{79.60}{0.67}
& \ExpRes{0.28}{0.01} \\

FS-EWCLoRA
& $\mathsf C+\mathsf F$
& $\mathsf C+\mathsf F+\mathsf P$
& \ExpRes{87.30}{0.57}
& \ExpRes{21.93}{2.98}
& \ExpRes{83.50}{0.50}
& \ExpRes{79.92}{1.04}
& \ExpRes{-3.80}{1.00}
& \ExpRes{81.71}{0.49}
& \ExpRes{0.43}{0.01} \\

Task-1 only
& $\mathsf C+\mathsf F$
& $\mathsf C+\mathsf P$
& \ExpRes{87.30}{0.57}
& \ExpRes{21.93}{2.98}
& \ExpRes{86.32}{0.86}
& \ExpRes{73.36}{1.20}
& \ExpRes{-0.98}{1.42}
& \ExpRes{79.84}{1.03}
& \ExpRes{0.30}{0.01} \\

Task-2 only
& $\mathsf C$
& $\mathsf C+\mathsf F+\mathsf P$
& \ExpRes{87.08}{0.38}
& \ExpRes{31.44}{4.24}
& \ExpRes{81.32}{0.50}
& \ExpRes{81.89}{1.04}
& \ExpRes{-5.76}{0.83}
& \ExpRes{81.61}{0.58}
& \ExpRes{0.47}{0.01} \\

\bottomrule
\end{tabular}
\end{table*}

\paragraph{Complementarity with FIM Representation}
\label{sec:complementarity}

% The preceding analysis shows that solution shaping changes the
% solution being protected rather than merely strengthening historical
% protection.
We next examine whether the performance gains from solution
shaping persist across different FIM schemes.
We evaluate six combinations of three FIM computation
schemes (batched empirical, per-example empirical, and
model-sampled) and two representations (diagonal and
Kronecker-factored).
% We now examine whether its benefit persists across
% different protection schemes, varying three FIM
% computations, including batched empirical, per-example empirical, and
% model-sampled, and two representations, including diagonal and
% Kronecker-factored.

Table~\ref{tab:complementarity} shows relatively modest performance
differences across the evaluated FIM computations and
representations.
FS-EWC improves FAA over the corresponding EWC
baseline by $6.08$ to $9.07$ percentage points across all six
configurations.
In these settings, the gains from FS-EWC exceed those obtained
by refining the FIM computation or representation alone.
Their consistency across protection schemes highlights the
complementary value of solution shaping: Fisher information
can guide the formation of solutions for future retention
as well as their subsequent protection.
See Appendix~\ref{app:complementarity_details} for $\lambda$
settings and sensitivity analyses.

\begin{table*}[t]
\centering
\captionsetup{font=footnotesize, labelfont=bf}
\caption{
\textbf{Performance under different Fisher structures
and computations.}
Under each structure, empirical FIM is computed using either
batch gradients (Batch) or per-sample gradients (Per-sample),
while Model denotes model-sampled FIM.
Each entry reports FAA / AAA (\%).
}
\label{tab:complementarity}

\footnotesize
\setlength{\tabcolsep}{3.8pt}
\renewcommand{\arraystretch}{1.05}

\begin{tabular*}{\textwidth}{
@{\extracolsep{\fill}}lcccccc@{}
}
\toprule

\multirow{2}{*}{{Method}}
& \multicolumn{3}{c}{{Diagonal}}
& \multicolumn{3}{c}{{Kronecker-factored}} \\
\cmidrule(lr){2-4}
\cmidrule(lr){5-7}

& {Batched}
& {Per-example}
& {Model-sampled}
& {Batched}
& {Per-example}
& {Model-sampled} \\

\midrule

EWC
& 51.96 / 66.30
& 50.60 / 66.60
& 50.76 / 66.60
& 53.74 / 67.37
& 52.75 / 67.55
& 52.71 / 67.33 \\

FS-EWC
& {59.29 / 70.69}
& {59.67/ 70.88}
& {59.62 / 70.86}
& {59.82 / 71.00}
& {59.95 / 71.16}
& {59.62 / 71.12} \\

\bottomrule
\end{tabular*}
\end{table*}

\begin{figure}[t]
\centering

\begin{minipage}[t]{0.48\linewidth}
\vspace{0pt}
\centering
\captionsetup{type=table,font=footnotesize,labelfont=bf}
\captionof{table}{
% Effect of retained-state construction on ImageNet-A.
% % Near-endpoint and trajectory-spanning averaging use learned
% % checkpoints from different portions of the optimization trajectory;
% Multi-seed results and experimental configurations are provided
% in Appendix~\ref{app:retained_state_details}.
\textbf{Constructing the anchor from learning trajectory.}
% Anchor construction in FS-EWC on ImageNet-A.
Anchor construction in FS-EWCLoRA on ImageNet-A, with current-task FIM
estimated at the resulting anchor.
Trajectory averaging achieves the highest mean AAA.
Multi-seed results and checkpoint configurations are provided in
Appendix~B.5.
% on ImageNet-A.
% The shaping objective and importance processing are fixed.
}
\label{tab:weight_merging_ablation}

\small
\setlength{\tabcolsep}{4pt}
\renewcommand{\arraystretch}{1.08}

\begin{tabular}{lccc}
\toprule
Retained state
& FAA$\uparrow$
& AAA$\uparrow$
& BWT$\uparrow$ \\
\midrule

% Endpoint
% & \textbf{59.97}
% & 71.00
% & $-8.66$ \\

% Near-endpoint averaging
% & 59.97
% & 70.97
% & $-7.93$ \\

% Trajectory averaging
% & 59.78
% & \textbf{71.29}
% & $-7.70$ \\

% \midrule

% Endpoint + start point
% & 58.92
% & 69.72
% & $-6.34$ \\

% Trajectory +  start point
% & 59.38
% & 70.94
% & $-7.58$ \\

Endpoint
& $59.23$
& $70.63$
& $-8.08$ \\

Late-stage
& $\mathbf{59.43}$
& $70.61$
& $-7.68$ \\

Trajectory
& $59.34$
& $\mathbf{70.84}$
& $-7.47$ \\

Dense trajectory
& $59.31$
& $70.745$
& $-7.44$ \\

\midrule

Endpoint + initial
& $58.44$
& $69.38$
& $\mathbf{-6.44}$ \\

Trajectory + initial
& $59.01$
& $70.59$
& $-7.33$ \\

\bottomrule
\end{tabular}
\label{table: merge}
\end{minipage}
\hfill
\begin{minipage}[t]{0.48\linewidth}
\vspace{0pt}
\centering
\includegraphics[width=\linewidth]
{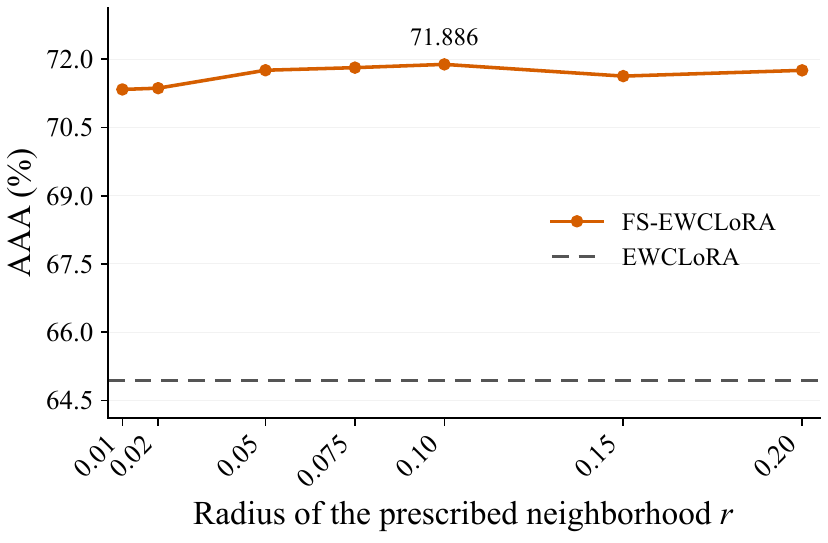}
\captionsetup{type=figure,font=footnotesize,labelfont=bf}
\captionof{figure}{
% Sensitivity to the prescribed neighborhood radius $r$ on ImageNet-A.
\textbf{Performance across neighborhood radii.}
AAA of FS-EWCLoRA on ImageNet-A as the prescribed neighborhood
radius $r$ varies.
% on ImageNet-A.
% The dashed line denotes EWCLoRA.
}
\label{fig:radius_sensitivity}
\end{minipage}

\end{figure}

% The contribution of Fisher sensitivity shaping has already been
% isolated .
% We focus on the second proposed component, retained-state
% merging, and on the prescribed neighborhood radius $r$.

\subsection{Ablation and Hyperparameter Studies}
Having isolated the roles of Fisher shaping and historical protection
in the preceding mechanism study, we further examine two implementation
choices of FS-EWC: retained-anchor construction and the prescribed
neighborhood radius.

% \minisection{Retained-state construction.}
% Table~\ref{tab:weight_merging_ablation} examines which optimization
% states should contribute to the retained anchor.
% Near-endpoint averaging leaves AAA essentially unchanged but
% improves BWT, indicating that nearby learned states still provide
% some benefit.
% Using checkpoints spanning the training trajectory further improves
% AAA and BWT, suggesting that the
% optimization trajectory contains additional information.
% In contrast, including the task initialization reduces FAA.
% Thus, the benefit does not arise simply from averaging more parameter
% states; rather, states formed along the optimization
% trajectory provide useful information for constructing the reference
% passed.

\paragraph{Retained-state construction.}
Table~\ref{tab:weight_merging_ablation} shows that trajectory averaging
achieves the highest AAA, while increasing checkpoint density provides
no further improvement.
Multi-checkpoint construction generally reduces forgetting relative to
the endpoint.
Including the initialization lowers FAA and AAA despite improving BWT.
These results suggest that the choice of retained state matters
beyond the training objective, but the benefit depends on which
learned states are combined rather than simply on their number.

\textbf{Neighborhood radius.}
Figure~\ref{fig:radius_sensitivity} evaluates the prescribed
neighborhood radius.
FS-EWCLoRA remains above the matched EWCLoRA baseline throughout the
tested range, while AAA varies only moderately across different
radii.
% The highest observed AAA is $71.886\%$ at $r=0.10$, with nearby
% settings yielding similar performance.
These results indicate that the benefit of sensitivity shaping is not
restricted to a narrowly tuned neighborhood radius.

\section{Conclusion}
EWC mitigates forgetting through Fisher-based parameter protection,
but remains constrained by the stability--plasticity trade-off.
Its historical protection is constructed from local parameter
importance, whereas forgetting under subsequent learning depends on
the sensitivity encountered after finite parameter displacement.
Motivated by this gap, we introduced Fisher-Shaped EWC (FS-EWC),
which incorporates Fisher information into current-task solution
formation while preserving subsequent
EWC protection. Our experiments show that shaped solutions can undergo larger subsequent displacement while incurring less forgetting, indicating a more favorable stability--plasticity balance. These findings suggest that future forgetting should be considered during learning.

\section*{AI use statement}

In this work, we used generative AI tools for assistance with mathematical formulation and proof development, methodological and experimental-design discussions, and interpretation of experimental results, and translation of manuscript content. We have not used generative AI tools to generate or modify experimental data, and tasks involving qualitative or thematic data analysis are not applicable to this work. Additionally, we used generative AI tools for manuscript organization, language editing, literature organization, reference formatting, and refinement of figures and captions.

We have reviewed all AI-assisted work. Mathematical statements and derivations were independently checked by the authors for correctness and consistency with the stated assumptions. Suggestions related to methodology, experimental design, and result interpretation were verified against the implemented methods, experimental protocols, and measured results. Literature-related content and references were checked against the original sources, and all AI-assisted manuscript text was manually reviewed and revised by the authors.

We take responsibility for the final content of this work, including text, claims, or artifacts produced with the aid of generative AI.

\subsection*{Ethics Statement}

This work studies algorithmic methods for replay-free continual learning and does not involve human subjects, personally identifiable information, or the collection of private data. All experiments are conducted on established public benchmarks. As with other machine learning methods, the behavior of the resulting models may inherit biases or limitations from the underlying datasets and pretrained models. Our work does not introduce additional data collection or deployment-specific risks beyond those associated with the datasets and model architectures considered.

\subsection*{Reproducibility Statement}

We provide the information necessary to reproduce both the theoretical and empirical results of this work. The FS-EWC objective, neighborhood-sensitivity optimization, retained-anchor construction, and Fisher-based consolidation procedure are specified in the main methodology section. Complete derivations and proofs of the theoretical results are provided in the appendix. The appendix further documents dataset and task construction, model architectures, continual-learning protocols, baseline configurations, optimization hyperparameters, Fisher estimation procedures, evaluation metrics, and all settings used in the ablation and diagnostic studies. Code and experiment configurations will be released to facilitate independent reproduction of the reported results.

% \subsection*{Reproducibility statement}

% (This section is \textbf{recommended} and does not count toward the page limit.)

% It is important that the work published in ICLR is reproducible. Authors are
% strongly encouraged to include a paragraph-long Reproducibility Statement at the
% end of the main text (before references) to discuss the efforts that have been
% made to ensure reproducibility. This paragraph should not itself describe
% details needed for reproducing the results, but rather reference the parts of
% the main paper, appendix, and supplemental materials that will help with
% reproducibility. For example, for novel models or algorithms, a link to an
% anonymous downloadable source code can be submitted as supplementary materials;
% for theoretical results, clear explanations of any assumptions and a complete
% proof of the claims can be included in the appendix; for any datasets used in
% the experiments, a complete description of the data processing steps can be
% provided in the supplementary materials. Each of the above are examples of
% things that can be referenced in the reproducibility statement.

% \subsubsection*{Author Contributions}
% If you'd like to, you may include  a section for author contributions as is done
% in many journals. This is optional and at the discretion of the authors.

% \subsubsection*{Acknowledgments}
% Use unnumbered third level headings for the acknowledgments. All
% acknowledgments, including those to funding agencies, go at the end of the paper.

\bibliography{iclr2027_conference}

@article{huszar2018note,
  title={Note on the quadratic penalties in elastic weight consolidation},
  author={Husz{\'a}r, Ferenc},
  journal={Proceedings of the National Academy of Sciences},
  volume={115},
  number={11},
  pages={E2496--E2497},
  year={2018},
  doi={10.1073/pnas.1717042115}
}

@inproceedings{schwarz2018progress,
  title={Progress \& Compress: A scalable framework for continual learning},
  author={Schwarz, Jonathan and Czarnecki, Wojciech and Luketina, Jelena and Grabska-Barwinska, Agnieszka and Teh, Yee Whye and Pascanu, Razvan and Hadsell, Raia},
  booktitle={Proceedings of the 35th International Conference on Machine Learning},
  series={Proceedings of Machine Learning Research},
  volume={80},
  pages={4528--4537},
  year={2018},
  publisher={PMLR}
}

@inproceedings{chaudhry2018riemannian,
  title={Riemannian Walk for Incremental Learning: Understanding Forgetting and Intransigence},
  author={Chaudhry, Arslan and Dokania, Puneet K. and Ajanthan, Thalaiyasingam and Torr, Philip H. S.},
  booktitle={Proceedings of the European Conference on Computer Vision},
  pages={532--547},
  year={2018}
}

@article{martens2020new,
  title={New insights and perspectives on the natural gradient method},
  author={Martens, James},
  journal={Journal of Machine Learning Research},
  volume={21},
  number={146},
  pages={1--76},
  year={2020}
}

@misc{vandeven2025computation,
  title={On the Computation of the Fisher Information in Continual Learning},
  author={van de Ven, Gido M.},
  year={2025},
  eprint={2502.11756},
  archivePrefix={arXiv},
  primaryClass={cs.LG}
}

@inproceedings{liu2018rotate,
  title={Rotate your Networks: Better Weight Consolidation and Less Catastrophic Forgetting},
  author={Liu, Xialei and Masana, Marc and Herranz, Luis and van de Weijer, Joost and Lopez, Antonio M. and Bagdanov, Andrew D.},
  booktitle={2018 24th International Conference on Pattern Recognition},
  pages={2262--2268},
  year={2018},
  organization={IEEE},
  doi={10.1109/ICPR.2018.8545895}
}

@inproceedings{martens2015kfac,
  title={Optimizing Neural Networks with Kronecker-factored Approximate Curvature},
  author={Martens, James and Grosse, Roger},
  booktitle={Proceedings of the 32nd International Conference on Machine Learning},
  series={Proceedings of Machine Learning Research},
  volume={37},
  pages={2408--2417},
  year={2015},
  publisher={PMLR}
}

@inproceedings{zheng2026ewclora,
title={Revisiting Weight Regularization for Low-Rank Continual Learning},
author={Yaoyue Zheng and Yin Zhang and Joost van de Weijer and Gido M van de Ven and Shaoyi Du and Xuetao Zhang and Zhiqiang Tian},
booktitle={The Fourteenth International Conference on Learning Representations},
year={2026},
url={https://openreview.net/forum?id=pZj2DhfaVD}
}

@InProceedings{SI2017Zenke,
  title = 	 {Continual Learning Through Synaptic Intelligence},
  author =       {Friedemann Zenke and Ben Poole and Surya Ganguli},
  booktitle = 	 {Proceedings of the 34th International Conference on Machine Learning},
  pages = 	 {3987--3995},
  year = 	 {2017},
  editor = 	 {Precup, Doina and Teh, Yee Whye},
  volume = 	 {70},
  series = 	 {Proceedings of Machine Learning Research},
  month = 	 {06--11 Aug},
  publisher =    {PMLR},
  url = 	 {https://proceedings.mlr.press/v70/zenke17a.html}
}

@inproceedings{liangInfLoRAInterferenceFreeLowRank2024a,
  title = {{{InfLoRA}}: {{Interference-Free Low-Rank Adaptation}} for {{Continual Learning}}},
  shorttitle = {{{InfLoRA}}},
  booktitle = {2024 {{IEEE}}/{{CVF Conference}} on {{Computer Vision}} and {{Pattern Recognition}} ({{CVPR}})},
  author = {Liang, Yan-Shuo and Li, Wu-Jun},
  date = {2024-06-16},
  pages = {23638--23647},
  publisher = {IEEE},
  location = {Seattle, WA, USA},
  doi = {10.1109/CVPR52733.2024.02231},
  url = {https://ieeexplore.ieee.org/document/10658274/},
  isbn = {979-8-3503-5300-6},
year={2024}
}

@article{krizhevsky2009learning,
  title={Learning multiple layers of features from tiny images},
  author={Krizhevsky, Alex and Hinton, Geoffrey and others},
  year={2009},
  publisher={Toronto, ON, Canada},
  journal={}
}

@InProceedings{Peng_2019_ICCV,
author = {Peng, Xingchao and Bai, Qinxun and Xia, Xide and Huang, Zijun and Saenko, Kate and Wang, Bo},
title = {Moment Matching for Multi-Source Domain Adaptation},
booktitle = {Proceedings of the IEEE/CVF International Conference on Computer Vision (ICCV)},
month = {October},
year = {2019}
}

@misc{meta2024llama32,
  author       = {{Meta}},
  title        = {{Llama 3.2} Model Card},
  year         = {2024},
  howpublished = {Model card},
  url          = {https://github.com/meta-llama/llama-models/blob/main/models/llama3_2/MODEL_CARD.md}
}

@inproceedings{wang2023orthogonal,
title={Orthogonal Subspace Learning for Language Model Continual Learning},
author={Xiao Wang and Tianze Chen and Qiming Ge and Han Xia and Rong Bao and Rui Zheng and Qi Zhang and Tao Gui and Xuanjing Huang},
booktitle={The 2023 Conference on Empirical Methods in Natural Language Processing},
year={2023},
url={https://openreview.net/forum?id=L7ZBpZZ8Va}
}

@inproceedings{NIPS2015_250cf8b5,
 author = {Zhang, Xiang and Zhao, Junbo and LeCun, Yann},
 booktitle = {Advances in Neural Information Processing Systems},
 pages = {},
 publisher = {Curran Associates, Inc.},
 title = {Character-level Convolutional Networks for Text Classification},
 url = {https://proceedings.neurips.cc/paper_files/paper/2015/file/250cf8b51c773f3f8dc8b4be867a9a02-Paper.pdf},
 volume = {28},
 year = {2015}
}

@misc{wu2025sdlorascalabledecoupledlowrank,
      title={{SD-LoRA}: Scalable Decoupled Low-Rank Adaptation for Class Incremental Learning}, 
      author={Yichen Wu and Hongming Piao and Long-Kai Huang and Renzhen Wang and Wanhua Li and Hanspeter Pfister and Deyu Meng and Kede Ma and Ying Wei},
      year={2025},
      eprint={2501.13198},
      archivePrefix={arXiv},
      primaryClass={cs.LG},
      url={https://arxiv.org/abs/2501.13198}, 
}

@InProceedings{ImagenetR_Hendrycks_2021_ICCV,
    author    = {Hendrycks, Dan and Basart, Steven and Mu, Norman and Kadavath, Saurav and Wang, Frank and Dorundo, Evan and Desai, Rahul and Zhu, Tyler and Parajuli, Samyak and Guo, Mike and Song, Dawn and Steinhardt, Jacob and Gilmer, Justin},
    title     = {The Many Faces of Robustness: A Critical Analysis of Out-of-Distribution Generalization},
    booktitle = {Proceedings of the IEEE/CVF International Conference on Computer Vision (ICCV)},
    month     = {October},
    year      = {2021},
    pages     = {8340-8349}
}

@article{dosovitskiy2020image,
  title={An image is worth 16x16 words: Transformers for image recognition at scale},
  author={Dosovitskiy, Alexey},
  journal={arXiv preprint arXiv:2010.11929},
  year={2020}
}

@article{EWC2017,
author = {James Kirkpatrick  and Razvan Pascanu  and Neil Rabinowitz  and Joel Veness  and Guillaume Desjardins  and Andrei A. Rusu  and Kieran Milan  and John Quan  and Tiago Ramalho  and Agnieszka Grabska-Barwinska  and Demis Hassabis  and Claudia Clopath  and Dharshan Kumaran  and Raia Hadsell },
title = {Overcoming catastrophic forgetting in neural networks},
journal = {Proceedings of the National Academy of Sciences},
volume = {114},
number = {13},
pages = {3521-3526},
year = {2017},
doi = {10.1073/pnas.1611835114},
URL = {https://www.pnas.org/doi/abs/10.1073/pnas.1611835114},}

@misc{maji2013finegrained,
      title={Fine-Grained Visual Classification of Aircraft}, 
      author={Subhransu Maji and Esa Rahtu and Juho Kannala and Matthew Blaschko and Andrea Vedaldi},
      year={2013},
      eprint={1306.5151},
      archivePrefix={arXiv},
      primaryClass={cs.CV},
      url={https://arxiv.org/abs/1306.5151}, 
}

@inproceedings{krause2013cars,
  title     = {3D Object Representations for Fine-Grained Categorization},
  author    = {Krause, Jonathan and Stark, Michael and Deng, Jia and Fei-Fei, Li},
  booktitle = {Proceedings of the IEEE International Conference on Computer Vision Workshops},
  pages     = {554--561},
  year      = {2013},
  doi       = {10.1109/ICCVW.2013.77}
}

@techreport{wah2011cub,
  title       = {The Caltech-UCSD Birds-200-2011 Dataset},
  author      = {Wah, Catherine and Branson, Steve and Welinder, Peter and Perona, Pietro and Belongie, Serge},
  institution = {California Institute of Technology},
  number      = {CNS-TR-2011-001},
  year        = {2011}
}

@inproceedings{nilsback2008automated,
  title     = {Automated Flower Classification over a Large Number of Classes},
  author    = {Nilsback, Maria-Elena and Zisserman, Andrew},
  booktitle = {2008 Sixth Indian Conference on Computer Vision, Graphics and Image Processing},
  pages     = {722--729},
  year      = {2008},
  publisher = {IEEE},
  doi       = {10.1109/ICVGIP.2008.47}
}

@inproceedings{parkhi2012cats,
  title     = {Cats and Dogs},
  author    = {Parkhi, Omkar M. and Vedaldi, Andrea and Zisserman, Andrew and Jawahar, C. V.},
  booktitle = {2012 IEEE Conference on Computer Vision and Pattern Recognition},
  pages     = {3498--3505},
  year      = {2012},
  publisher = {IEEE},
  doi       = {10.1109/CVPR.2012.6248092}
}

@InProceedings{imagenet-a,
    author    = {Hendrycks, Dan and Zhao, Kevin and Basart, Steven and Steinhardt, Jacob and Song, Dawn},
    title     = {Natural Adversarial Examples},
    booktitle = {Proceedings of the IEEE/CVF Conference on Computer Vision and Pattern Recognition (CVPR)},
    month     = {June},
    year      = {2021},
    pages     = {15262-15271}
}

@ARTICLE{Wang2024survey,
  author={Wang, Liyuan and Zhang, Xingxing and Su, Hang and Zhu, Jun},
  journal={IEEE Transactions on Pattern Analysis and Machine Intelligence}, 
  title={A Comprehensive Survey of Continual Learning: Theory, Method and Application}, 
  year={2024},
  volume={46},
  number={8},
  pages={5362-5383},
  doi={10.1109/TPAMI.2024.3367329}}

@article{Shaheen2022,
  author  = {Shaheen, Khadija and Hanif, Muhammad Abdullah and Hasan, Osman and Shafique, Muhammad},
  title   = {Continual Learning for Real-World Autonomous Systems: Algorithms, Challenges and Frameworks},
  journal = {Journal of Intelligent \& Robotic Systems},
  year    = {2022},
  volume  = {105},
  number  = {1},
  pages   = {9},
  doi     = {10.1007/s10846-022-01603-6},
  url     = {https://doi.org/10.1007/s10846-022-01603-6},
  issn    = {1573-0409}
}

@inproceedings{ICLR2026_9c84feb7,
 author = {Liu, Yuhang and Li, Tao and Huang, Zhehao and Yang, Zuopeng and Huang, Xiaolin},
 booktitle = {International Conference on Learning Representations},
 editor = {C. Vondrick and B. Hariharan and C. Raffel and L. Pinto and D. Yang and A. Faust},
 pages = {96782--96813},
 title = {Bi-LoRA: Efficient Sharpness-Aware Minimization for Fine-Tuning Large-Scale Models},
 url = {https://proceedings.iclr.cc/paper_files/paper/2026/file/9c84feb75eae1ef6389f31b3ef050b6a-Paper-Conference.pdf},
 volume = {2026},
 year = {2026}
}

@INPROCEEDINGS{CLLORA,
  author={He, Jiangpeng and Duan, Zhihao and Zhu, Fengqing},
  booktitle={2025 IEEE/CVF Conference on Computer Vision and Pattern Recognition (CVPR)}, 
  title={CL-LoRA: Continual Low-Rank Adaptation for Rehearsal-Free Class-Incremental Learning}, 
  year={2025},
  volume={},
  number={},
  pages={30534-30544},
  doi={10.1109/CVPR52734.2025.02843}}

@misc{qian2026treeloraefficientcontinuallearning,
      title={TreeLoRA: Efficient Continual Learning via Layer-Wise LoRAs Guided by a Hierarchical Gradient-Similarity Tree}, 
      author={Yu-Yang Qian and Yuan-Ze Xu and Zhen-Yu Zhang and Peng Zhao and Zhi-Hua Zhou},
      year={2026},
      eprint={2506.10355},
      archivePrefix={arXiv},
      primaryClass={cs.LG},
      url={https://arxiv.org/abs/2506.10355}, 
}
\bibliographystyle{iclr2027_conference}

\newpage
\appendix
% Replace the existing \section{Appendix} and its contents with this block.
% Keep the manuscript's existing \appendix command and \end{document}.
% The final subsection condenses the separately supplied retention appendix.
% \section{Appendix I - Theory}
\section{Supplementary Theoretical Analysis}
\label{app:theory}

\subsection{Notation}
\label{app:notation}

Table~\ref{tab:main_notation} collects the notation used below.
All gradients, perturbations, and importance matrices use the same
trainable parameterization.

\begin{table}[htbp]
\centering
\caption{Key notation used in FS-EWC.}
\label{tab:main_notation}
\small
\renewcommand{\arraystretch}{1.1}
\setlength{\tabcolsep}{5pt}
\begin{tabular}{@{}p{0.32\linewidth}p{0.64\linewidth}@{}}
\toprule
\textbf{Symbol} & \textbf{Meaning} \\
\midrule
$\vtheta$, $\vtheta_t$ & Candidate parameters and the retained task-$t$ anchor. \\
$\mF_t^{\mathrm{emp}}$, $\widehat{\mF}_t$, $\varepsilon_F$  & Empirical FIM and the normalized and damped current-task FIM,  Fisher-normalization stabilizer.  \\
$\mLambda_t$, $\Omega_{t-1}$ & Consolidated importance and the inherited EWC penalty. \\
$\vDelta$, $\vepsilon$ & Future parameter displacement and sensitivity-search perturbation. \\
$D_t(\vtheta,\vDelta)$ & Task-loss change; positive values indicate deterioration. \\
$\mathcal R_{\mathrm F,t}$, $\mathcal R_{\mathrm F,t}^{\max}$ & Fisher trace and its maximum over $\sU_r(\vtheta)$. \\
$\vg$, $\mH$ & Gradient and Hessian of the mean loss on a fixed batch $\sB$. \\
$s(\vu)$, $\widehat{\vh}(\vu)$ & Probe step size and estimate of $2\mH(\vu)\vg(\vu)$. \\
$\vz$, $\vd_{t,\sB}$ & Selected neighborhood point and numerical update direction. \\
$\sK_t$, $K$, $\overline{\vtheta}_t$ & Selected checkpoint indices, their number, and their parameter average. \\
$\beta$, $\gamma$, $\lambda$ & Damping, decay coefficient, and historical-regularization. \\
$r$,$\rho$, $\xi$ & Neighborhood radius, finite-difference probe scale. and normalization stabilizer. \\
% $\rho$, $\varepsilon$, $\underline r_{t,\varepsilon}(r)$ & Admissible drift radius, loss tolerance, and certified retention radius. \\
\bottomrule
\end{tabular}
\end{table}

\subsection{Evaluation quantities.}
\label{app:diagnostic_metrics}

Let $a_{t,i}$ denote the test accuracy on Task~$i$ after learning
through Task~$t$, where $1\leq i\leq t\leq T$.
The average accuracy after learning $t$ tasks is
\begin{equation}
A_t
=
\frac{1}{t}
\sum_{i=1}^{t} a_{t,i}.
\label{eq:stage_average_accuracy}
% \notag
\end{equation}
This is the stage-wise average accuracy reported in the learning
curves.
We summarize continual-learning performance using final average
accuracy (FAA), average accuracy across learning stages (AAA), and
backward transfer (BWT):
\begin{equation}
\mathrm{FAA}
=
A_T,\quad
% =
% \frac{1}{T}
% \sum_{i=1}^{T}a_{T,i},
\mathrm{AAA}
=
\frac{1}{T}
\sum_{t=1}^{T}A_t
% =
% \frac{1}{T}
% \sum_{t=1}^{T}
% \frac{1}{t}
% \sum_{i=1}^{t}a_{t,i},
\quad
\mathrm{BWT}
=
\frac{1}{T-1}
\sum_{i=1}^{T-1}
\left(
a_{T,i}-a_{i,i}
\right).
% \notag
\end{equation}
A more negative BWT indicates greater forgetting.
% All task averages weight the observed tasks equally

% Let $a_{t,i}$ denote the test accuracy on Task~$i$ after learning
% through Task~$t$, where $1\leq i\leq t\leq T$.
% We use three standard metrics to summarize continual-learning
% performance.
% Final average accuracy (FAA) measures the average performance on all
% tasks after the complete sequence:
% \begin{equation}
% \mathrm{FAA}
% =
% \frac{1}{T}
% \sum_{i=1}^{T} a_{T,i}.
% \end{equation}
% Average accuracy across learning stages (AAA) additionally accounts
% for performance throughout the learning process:
% \begin{equation}
% \mathrm{AAA}
% =
% \frac{1}{T}
% \sum_{t=1}^{T}
% \left(
% \frac{1}{t}
% \sum_{i=1}^{t} a_{t,i}
% \right).
% \end{equation}
% Backward transfer (BWT) measures the final change in previously
% learned tasks relative to their performance when first learned:
% \begin{equation}
% \mathrm{BWT}
% =
% \frac{1}{T-1}
% \sum_{i=1}^{T-1}
% \left(
% a_{T,i}-a_{i,i}
% \right).
% \end{equation}
% A more negative BWT indicates greater forgetting.
% All task averages weight observed tasks equally.

For the solution-shaping diagnostics, we additionally characterize
parameter displacement and solution sensitivity in the effective LoRA update space.
Specifically, we write
\begin{equation}
\mathbf w_t
=
\operatorname{vec}
\left(
\left\{
B_{\ell c,t}A_{\ell c,t}
\right\}_{\ell,\;c\in\{Q,V\}}
\right)
\end{equation}
for the vectorized effective LoRA weights retained after Task~$t$.
For realized subsequent learning, define
\begin{equation}
\Delta\mathbf w_t
=
\mathbf w_t-\mathbf w_1,
\qquad
d_t
=
\|\Delta\mathbf w_t\|_2,
\end{equation}
and the signed Task~1 test-loss change
\begin{equation}
D_{1,t}^{\mathrm{test}}
=
\widehat{\mathcal L}_{1,\mathrm{test}}(\mathbf w_t)
-
\widehat{\mathcal L}_{1,\mathrm{test}}(\mathbf w_1).
\end{equation}
For $d_t>0$, let
$\mathbf v_t=\Delta\mathbf w_t/d_t$.
The displacement-normalized directional Fisher sensitivity along the
straight segment connecting the two endpoints is
\begin{equation}
\overline{\mathcal S}_{\mathrm F,1}^{\mathrm{path}}(t)
=
\int_0^1
\mathbf v_t^\top
\mF_{1,\mathrm{tr}}^{\mathrm{emp}}
(\mathbf w_1+s\Delta\mathbf w_t)
\mathbf v_t\,ds .
\end{equation}

For the controlled-neighborhood diagnostic, let $d$ denote the
prescribed perturbation radius and $\mathbf u_k$ a unit perturbation
direction.
The Task~1 training-loss change along direction $k$ is
\begin{equation}
D_{1,k}^{\mathrm{nbr}}(d)
=
\widehat{\mathcal L}_{1,\mathrm{tr}}
(\mathbf w_1+d\mathbf u_k)
-
\widehat{\mathcal L}_{1,\mathrm{tr}}(\mathbf w_1).
\end{equation}
We report
\begin{equation}
\overline D_1^{+}(d)
=
\frac{1}{K_D}
\sum_{k=1}^{K_D}
[D_{1,k}^{\mathrm{nbr}}(d)]_+,
\qquad
\overline{\mathcal R}_{\mathrm F,1}^{\mathrm{nbr}}(d)
=
\frac{1}{K_F}
\sum_{k=1}^{K_F}
\tr
\mF_{1,\mathrm{tr}}^{\mathrm{emp}}
(\mathbf w_1+d\mathbf u_k).
\end{equation}
For an allowed mean positive loss increase $\tau$, the measured
tolerance radius over the evaluated grid
$0=d_0<d_1<\cdots<d_J$ is
\begin{equation}
\widehat d_\tau
=
\max
\left\{
d_j:
\overline D_1^{+}(d_i)\leq\tau
\ \text{for all } i\leq j
\right\}.
\end{equation}
Because $\widehat d_\tau$ is evaluated only on the measured grid,
it is an empirical summary rather than a certificate over the
continuous neighborhood.

For the two-task diagnostic, $A_2$ is the corresponding
stage-average accuracy from Eq.~\eqref{eq:stage_average_accuracy},
i.e., $A_2=(a_{21}+a_{22})/2$.
In this case, $\mathrm{BWT}=a_{21}-a_{11}$ and
$d_2=\|\mathbf w_2-\mathbf w_1\|_2$.

% For the two-task analysis, let $a_{ij}$ denote the test accuracy on
% Task~$j$ after learning through Task~$i$.
% We therefore report
% \begin{equation}
% \mathrm{BWT}=a_{21}-a_{11},
% \qquad
% A_2=\frac{a_{21}+a_{22}}{2},
% \qquad
% d_2=\|\mathbf w_2-\mathbf w_1\|_2.
% \end{equation}
% We additionally denote by $\mathcal B_1$ the Task~1 endpoint value of
% the implementation-level batch-gradient Fisher proxy.
% Unlike
% $\overline{\mathcal R}_{\mathrm F,1}^{\mathrm{nbr}}(d)$,
% $\mathcal B_1$ is not a per-example empirical-Fisher trace and the two
% quantities should not be compared numerically.

\subsection{Proof of Finite-Displacement Sensitivity}
\label{app:finite_displacement_fisher}

Fix $\|\vDelta\|_2\leq r$. The segment
$\vtheta+\tau\vDelta$, $\tau\in[0,1]$, lies in $\sU_r(\vtheta)$,
so the differentiability assumption in
Proposition~\ref{prop:finite_displacement_fisher} gives
\begin{equation}
D_t(\vtheta,\vDelta)
=\int_0^1\frac1{m_t}\sum_{i=1}^{m_t}
\vg_{t,i}(\vtheta+\tau\vDelta)^\top\vDelta\,\mathrm d\tau.
\label{eq:app_task_loss_integral}
\end{equation}
Applying Cauchy--Schwarz to the integral and then to the sample
average yields
\begin{equation}
\begin{aligned}
|D_t(\vtheta,\vDelta)|^2
&\leq\int_0^1\left[
\frac1{m_t}\sum_{i=1}^{m_t}
\vg_{t,i}(\vtheta+\tau\vDelta)^\top\vDelta
\right]^2\mathrm d\tau
\\
&\leq\int_0^1\frac1{m_t}\sum_{i=1}^{m_t}
\left[\vg_{t,i}(\vtheta+\tau\vDelta)^\top\vDelta\right]^2
\mathrm d\tau
\\
&=\int_0^1\vDelta^\top
\mF_t^{\mathrm{emp}}(\vtheta+\tau\vDelta)\vDelta\,\mathrm d\tau.
\end{aligned}
\label{eq:app_finite_displacement_conclusion}
\end{equation}
This proves the proposition.
% The argument requires neither a stationary
% anchor nor a Fisher--Hessian identification; the integration segment
% need not coincide with the subsequent optimization trajectory.

\subsection{FS-EWC from a Bayesian Perspective}
\label{app:bayesian_ewc_surrogates}

The Bayesian perspective provides a complementary interpretation of
both components of FS-EWC: accounting for a parameter distribution
when learning its center, and selecting the center retained for
subsequent tasks.

\paragraph{EWC from a Bayesian perspective.}
Sequential Bayesian learning maintains a distribution over model
parameters. For conditionally independent task data,
\begin{equation}
\begin{aligned}
p(\vtheta\mid\sS_{1:t})
\propto
p(\vtheta\mid\sS_{1:t-1})
p(\sS_t\mid\vtheta).
\end{aligned}
\label{eq:app_bayes_recursive_update}
% \notag
\end{equation}
Thus, the posterior after task $t-1$ serves as the prior for
learning task $t$, carrying historical information without
requiring the original data.

The exact posterior is generally intractable for neural networks.
EWC  approximates the inherited parameter distribution by
a local Gaussian surrogate
\citep{EWC2017,huszar2018note}:
\begin{equation}
q_{t-1}(\vtheta)
=
\mathcal N
\left(
\vtheta;
\vtheta_{t-1},
\mLambda_{t-1}^{-1}
\right) .
\label{eq:app_bayes_inherited_gaussian}
\end{equation}
Using this surrogate as the prior, a MAP-style current-task update is
\begin{equation}
\begin{aligned}
\vtheta_t^{\mathrm{MAP}}
=
\arg\min_{\vtheta}
\left[
\widehat{\mathcal L}_t^{\mathrm{CE}}(\vtheta)
+
\frac{\lambda}{2}
(\vtheta-\vtheta_{t-1})^\top
\mLambda_{t-1}
(\vtheta-\vtheta_{t-1})
\right] = \mathcal J_t^{\mathrm{EWC}}(\vtheta),
\end{aligned}
\label{eq:app_bayes_ewc_map}
\end{equation}
where $\lambda$ absorbs the practical scaling between the mean
current-task loss and the inherited precision.
This recovers the EWC objective in
~\eqref{eq:online_ewc_objective}.
Hence, EWC retains a distribution-derived representation of
historical uncertainty, but performs current-task learning through
a point estimate of the new parameters.

\paragraph{Optimizing the distribution.}
The underlying Bayesian update is distributional rather than
pointwise. A corresponding variational objective is
\begin{equation}
\mathcal V_t(Q)
=
\mathbb E_{\vu\sim Q}
[
\widehat{\mathcal L}_t^{\mathrm{CE}}(\vu)
]
+
\lambda\mathrm{KL}(Q\|q_{t-1}),
\label{eq:app_bayes_practical_variational_update}
\end{equation}
where $Q$ is a candidate current-task parameter distribution.
The first term evaluates current-task performance across parameter
values represented by $Q$, whereas standard EWC evaluates the
current-task loss only at its MAP candidate.

To expose this difference, consider a fixed-shape local family
$Q_{\vtheta}$ represented by
\begin{equation}
\vu=\vtheta+\vepsilon,
\qquad
\mathbb E[\vepsilon]=\bm0,
\qquad
\|\vepsilon\|_2\leq r.
\end{equation}
Its expected current-task loss decomposes as
\begin{equation}
\begin{aligned}
\mathbb E_{\vu\sim Q_{\vtheta}}
[
\widehat{\mathcal L}_t^{\mathrm{CE}}(\vu)
]
=
\widehat{\mathcal L}_t^{\mathrm{CE}}(\vtheta)
+
\mathbb E[
D_t(\vtheta,\vepsilon)
].
\end{aligned}
\label{eq:app_bayes_distribution_averaged_loss}
\end{equation}
The second term measures the loss variation over parameter values
around the candidate center and is not controlled by center loss alone.

For this fixed-shape, zero-mean family, the historical KL term has
the same center-dependent part as EWC:
\begin{equation}
\lambda\mathrm{KL}(Q_{\vtheta}\|q_{t-1})
=
\Omega_{t-1}(\vtheta)+C,
\label{eq:app_bayes_location_kl}
\end{equation}
where $C$ does not depend on $\vtheta$.
Using the finite-displacement Fisher bound,
\begin{equation}
\mathbb E[
D_t(\vtheta,\vepsilon)
]
\leq
r\sqrt{
\mathcal R_{\mathrm F,t}^{\max}(\vtheta;r)
}
\leq
\mathcal R_{\mathrm F,t}^{\max}(\vtheta;r)
+
\frac{r^2}{4}.
\end{equation}
Therefore,
\begin{equation}
\mathcal V_t(Q_{\vtheta})
\leq
\mathcal J_t^{\mathrm {FS-EWC}}(\vtheta;r)
+
C+\frac{r^2}{4}.
\label{eq:app_bayes_same_training_objective}
\end{equation}

For fixed distribution shape, $r$, the
parameter-dependent upper surrogate is exactly the FS-EWC objective.
Thus, FS-EWC can be viewed as extending EWC's pointwise MAP update
toward distribution-aware center selection: the center is chosen not
only for its own task loss and historical compatibility, but also for
controlled loss variation over its surrounding parameter distribution.

\paragraph{Selecting the retained center.}
At task completion, let $Q_t^{\mathrm{traj}}$ assign equal mass to
the selected late-stage checkpoints $\vtheta_{t,k}$, $k\in\sK_t$.
For any fixed positive definite matrix $\mA$, fitting a Gaussian
center to this empirical distribution yields
\begin{equation}
\begin{aligned}
\arg\min_{\vtheta}
\mathbb E_{\vu\sim Q_t^{\mathrm{traj}}}
[-\log\mathcal N(\vu;\vtheta,\mA^{-1})]
\quad=\frac1K\sum_{k\in\sK_t}\vtheta_{t,k}
=\overline{\vtheta}_t.
\end{aligned}
\label{eq:app_bayes_anchor_selection}
\end{equation}
The equality follows by minimizing the associated quadratic in the
center and recovers the retained-anchor rule
$\vtheta_t:=\overline{\vtheta}_t$.
It treats the selected checkpoints as an empirical distribution,
without assuming they are exact posterior samples.
Fisher importance is then estimated at the retained anchor and
consolidated as specified in Section~\ref{sec:fs_consolidation}.
Thus, the distributional view motivates both learning the center
through Fisher shaping and selecting the retained center through
trajectory averaging.

\subsection{Practical Implementation}
\label{app:fisher_numerical_details}
\label{sec:fs_optimization}

% This appendix presents the practical solver for the batch objective and briefly
% discusses its two numerical approximations: finite-difference
% sensitivity estimation and single-step neighborhood search.
% The separate replacement of the empirical-Fisher trace by the
% squared batch gradient is discussed in
% Appendix~\ref{app:fisher_gradient_surrogate}.
\subsubsection{Empirical Fisher Computation}
\label{app:fisher_gradient_surrogate}

Several Fisher variants are used in continual learning, differing in
the distribution over labels and in how expectations are approximated.
Our theoretical development uses the per-example empirical Fisher in
Eq.~\eqref{eq:empirical_fisher_definition}. We focus here on the
specific approximation used by our method and refer to
\citet{martens2020new,vandeven2025computation} for broader treatments
of Fisher definitions and computation.

The theoretical sensitivity target is defined using the
per-example empirical-Fisher trace, whereas practical
mini-batch optimization uses the squared gradient of the
batch-averaged loss.
For a fixed batch $\sB$ with $b=|\sB|$, let
\[
\vg(\vu)
=
\frac1b\sum_{i\in\sB}\vg_{t,i}(\vu).
\]
The two quantities satisfy the exact decomposition
\begin{equation}
\begin{aligned}
\frac1b\sum_{i\in\sB}\|\vg_{t,i}(\vu)\|_2^2
=
\|\vg(\vu)\|_2^2
+
\frac1b\sum_{i\in\sB}
\|\vg_{t,i}(\vu)-\vg(\vu)\|_2^2.
\end{aligned}
\label{eq:app_fs_gradient_decomposition}
\end{equation}
Thus, the squared batch gradient omits the nonnegative
dispersion of per-example gradients and is generally not
identical to the empirical-Fisher trace.
We use it as a computational surrogate because the
batch-averaged gradient is directly available during
stochastic optimization.

\subsubsection{Fisher Post-processing and Consolidation}
% \label{app:fisher_normalization}
\label{app:fisher_normalization}

Following prior work \citep{schwarz2018progress} on Fisher normalization,
we normalize the current-task Fisher importance to reduce arbitrary
task-dependent differences in its absolute scale and make consolidation
depend primarily on the relative importance of protected parameters.
Specifically, we use global trace normalization over all protected
parameter coordinates:
\begin{equation}
\widetilde{\mF}_t
=
\mathcal N\!\left(
\Diag\!\left(\mF_t^{\mathrm{emp}}(\vtheta_t)\right)
\right)
=
\frac{
\Diag\!\left(\mF_t^{\mathrm{emp}}(\vtheta_t)\right)
}{
\operatorname{tr}\!\left(
\Diag\!\left(\mF_t^{\mathrm{emp}}(\vtheta_t)\right)
\right)
+\xi_F
},
\qquad
\xi_F=10^{-12}.
\label{eq:fisher_trace_normalization}
\end{equation}
The trace is computed jointly over all protected coordinates rather
than independently for individual layers. For LoRA adaptation, these
coordinates correspond to the effective weight updates $\Delta W$.
Because all Fisher entries share the same normalization factor, their
relative importance is preserved, while for nondegenerate Fisher
statistics,
$\operatorname{tr}(\widetilde{\mF}_t)\approx1$.

We then apply isotropic damping,
\begin{equation}
\widehat{\mF}_t
=
(1-\beta)\widetilde{\mF}_t
+
\beta\frac{\mI}{d},
\qquad
\beta\in[0,1],
\label{eq:fisher_damping}
\end{equation}
where $d$ is the total number of protected coordinates and
$\beta=0.2$ by default.
Since $\operatorname{tr}(\mI/d)=1$, damping preserves the normalized
trace, i.e.,
$\operatorname{tr}(\widehat{\mF}_t)\approx1$.

Following recursive consolidation as in online EWC \citep{schwarz2018progress}, we update
\begin{equation}
\mLambda_t
=
\gamma\mLambda_{t-1}
+
\widehat{\mF}_t,
\qquad
\gamma\in[0,1],
\label{eq:fisher_recursive_consolidation}
\end{equation}
with $\gamma=1$ by default.

\subsubsection{Practical Sensitivity Solver}
\label{app:fs_numerical_update}

For a fixed task $t$ and training batch $\sB$, let
\begin{equation}
\begin{aligned}
\vg(\vu)
=
\nabla_{\vu}
\widehat{\mathcal L}_{t,\sB}^{\mathrm{CE}}(\vu), \qquad
\mH(\vu)
=
\nabla_{\vu}^{2}
\widehat{\mathcal L}_{t,\sB}^{\mathrm{CE}}(\vu).
\end{aligned}
\label{eq:batch_loss_gradient_hessian}
\notag
\end{equation}
For twice continuously differentiable batch losses,
the derivative of the squared-gradient is
\begin{equation}
\nabla_{\vu}\|\vg(\vu)\|_2^2
=
2\mH(\vu)\vg(\vu).
\label{eq:squared_sensitivity_gradient}
\end{equation}

% \paragraph{Sensitivity-gradient estimator.}
\paragraph{Sensitivity-gradient estimator.}
We approximate this derivative using a normalized finite difference
probe with a fixed unit scale. At a generic parameter point $\vu$,
define
\begin{equation}
\begin{aligned}
\mathbf{s}(\vu)
&=
\frac{\vg(\vu)}
{\|\vg(\vu)\|_2+\xi},
\\
\widehat{\vh}(\vu)
&=
2\bigl(\|\vg(\vu)\|_2+\xi\bigr)
\left[
\vg\bigl(\vu+\rho \mathbf{s}(\vu)\bigr)
-
\vg(\vu)
\right]/\rho.
\end{aligned}
\label{eq:squared_fisher_gradient_estimator}
\end{equation}
Here, $\mathbf{s}(\vu)$ specifies the normalized probe direction,
$\rho$ controls the finite-difference probe scale, and
$\xi>0$ stabilizes the normalization.
Since
$\|\rho\mathbf{s}(\vu)\|_2\leq\rho$,
$\rho$ determines the maximum probe displacement.
Unless otherwise specified, we set $\rho=1$.
% Here, $\mathbf{s}(\vu)$ is the normalized finite difference probe and
% $\xi>0$ stabilizes the normalization.
% The probe scale is fixed to 1
% throughout and is not treated as a tunable hyperparameter.
When the Hessian varies slowly along the probe segment,
\begin{equation}
\widehat{\vh}(\vu)
\approx
2\mH(\vu)\vg(\vu)
=
\nabla_{\vu}\|\vg(\vu)\|_2^2.
\label{eq:squared_fisher_gradient_estimator_target}
\end{equation}

% We estimate this derivative using a normalized finite-difference
% probe. At a generic parameter point $\vu$, define
% \begin{equation}
% \begin{aligned}
% \mathbf{s}(\vu)
% &=
% \frac{\vg(\vu)}
% {\|\vg(\vu)\|_2+\xi},
% \\
% \widehat{\vh}(\vu)
% &=
% \frac{2(\|\vg(\vu)\|_2+\xi)}{r}
% \left[
% \vg\bigl(\vu+\mathbf{s}(\vu)\bigr)
% -
% \vg(\vu)
% \right].
% \end{aligned}
% \label{eq:squared_fisher_gradient_estimator}
% \end{equation}
% Here, $\mathbf{s}(\vu)$ is the finite-difference probe displacement,
% $r>0$ controls its maximum length, and $\xi>0$ stabilizes
% normalization.
% Locally,
% \begin{equation}
% \widehat{\vh}(\vu)
% \approx
% 2\mH(\vu)\vg(\vu)
% =
% \nabla_{\vu}\|\vg(\vu)\|_2^2.
% \label{eq:squared_fisher_gradient_estimator_target}
% \notag
% \end{equation}
\paragraph{Single-step neighborhood search and update.}
We approximate the inner maximization by a single normalized ascent
step. The estimated sensitivity gradient determines the search
direction, while the prescribed radius $r$ controls the extent of
the neighborhood search:
\begin{equation}
\begin{aligned}
\widehat{\vepsilon}(\vtheta)
&=
r
\frac{\widehat{\vh}(\vtheta)}
{\|\widehat{\vh}(\vtheta)\|_2+\xi},
\\
\vz
&=
\vtheta+\widehat{\vepsilon}(\vtheta),
\\
\vd_{t,\sB}
&=
\vg(\vtheta)
+
\widehat{\vh}(\vz)
+
\lambda\mLambda_{t-1}
(\vtheta-\vtheta_{t-1}).
\end{aligned}
\label{eq:squared_fisher_numerical_update}
\end{equation}
By construction,
$\|\widehat{\vepsilon}(\vtheta)\|_2\leq r$, so the selected point
$\vz$ remains within the prescribed neighborhood of $\vtheta$.
The first estimator call provides a first-order approximation to the inner search direction used to
select this neighborhood point, while the second approximates the
sensitivity gradient at $\vz$ and supplies the outer shaping
correction.

% \paragraph{Neighborhood search and update.}
% The sensitivity estimate at the current parameters determines a
% single ascent step within the prescribed neighborhood.
% The same estimator evaluated at the resulting point provides the
% outer sensitivity correction:
% \begin{equation}
% \begin{aligned}
% \vz
% &=
% \vtheta
% + \widehat{\vh}(\vtheta),
% % \frac{\widehat{\vh}(\vtheta)}
% % {\|\widehat{\vh}(\vtheta)\|_2+\xi},
% \\
% \vd_{t,\sB}
% &=
% \vg(\vtheta)
% +
% \widehat{\vh}(\vz)
% +
% \lambda\mLambda_{t-1}
% (\vtheta-\vtheta_{t-1}).
% \end{aligned}
% \label{eq:squared_fisher_numerical_update}
% \end{equation}
% The first estimator call selects a high-sensitivity neighborhood
% location, while the second supplies the sensitivity correction at
% that location.

% All finite-difference evaluations within one update use the same
% mini-batch and stochastic forward-pass realization.
% Probe evaluations do not update optimizer state or persistent model
% state.
% Each estimator uses two loss-gradient evaluations.
% % , resulting in four
% % current-task gradient evaluations per update
% The initial gradient is reused for current-task fitting.
% The base optimizer is always applied to the original parameters
% $\vtheta$, and the inherited EWC penalty is added only to the final
% update direction.
% Before consolidation, we normalize the current-task Fisher importance
% to reduvariation. This makes consolidation
% depend primarily on the relative importance of parameters within each
% task, rather than on arbitrary differences in the absolute Fisher
% magnitude across tasks.

\subsubsection{Complete Learning Procedure}
\label{app:fs_learning_procedure}

Algorithm~\ref{alg:fs_ewc} summarizes the complete FS-EWC procedure,
including within-task sensitivity shaping, retained-anchor
construction, and task-boundary Fisher consolidation.

\begin{algorithm}[h]
\caption{Fisher-Shaped EWC (FS-EWC)}
\label{alg:fs_ewc}
\begin{algorithmic}[1]

\REQUIRE Initial parameters $\vtheta_0$;
datasets $\sS_1,\ldots,\sS_T$;
$\lambda,\beta,r,\rho,\gamma$;
checkpoint sets $\sK_t$;
base optimizer.

\STATE Initialize $\mLambda_0\leftarrow\mzero$.

\FOR{$t=1,\ldots,T$}

    \STATE Set $\vtheta\leftarrow\vtheta_{t-1}$ and reset the running
    checkpoint average.

    \FOR{each training batch $\sB$ from $\sS_t$}

        \STATE Evaluate $\widehat{\vh}(\vtheta)$ using
        Eq.~\ref{eq:squared_fisher_gradient_estimator}.

        \STATE Construct the neighborhood point $\vz$ using
        Eq.~\ref{eq:squared_fisher_numerical_update}.

        \STATE Evaluate $\widehat{\vh}(\vz)$ using the same batch.

        \STATE Form $\vd_{t,\sB}$ using
        Eq.~\ref{eq:squared_fisher_numerical_update}.

        \STATE Update the unperturbed parameters $\vtheta$ with the
        base optimizer using $\vd_{t,\sB}$.

        \STATE Accumulate $\vtheta$ when the current checkpoint belongs
        to $\sK_t$.

    \ENDFOR

    \STATE Construct the retained anchor
    $\vtheta_t$ using
    Eq.~\ref{eq:averaged_task_anchor}.

    \STATE Estimate
$\mF_t^{\mathrm{emp}}(\vtheta_t)$
using the current-task data.

\STATE Form the normalized and damped current-task importance
$\widehat{\mF}_t$ using
Eqs.~\ref{eq:fisher_trace_normalization}
and~\ref{eq:fisher_damping}.

\STATE Consolidate historical importance:
$\mLambda_t
\leftarrow
\gamma\mLambda_{t-1}
+
\widehat{\mF}_t$
using Eq.~\ref{eq:fisher_recursive_consolidation}.

%     \STATE Estimate $\widehat{\mF}_t$ at $\vtheta_t$ using the
%     current-task data.
%     \STATE  $\mLambda_t
%     \leftarrow \widehat{\mF}_t$ Normalize, damp, and consolidate the FIM
% following Sec.~\ref{app:fisher_normalization}.

%     \STATE Add damp term and normilizing the fisher $
% \widehat{\mF}_t
%   \leftarrow
% \mathcal N\!\left(
% \Diag\!\left(\mF_t^{\mathrm{emp}}(\vtheta_t)\right)
% \right)
% +\beta\mI,
% \qquad \beta>0,
% $

%     \STATE Update
%     $\mLambda_t
%     \leftarrow
%     \gamma\mLambda_{t-1}
%     +
%     \widehat{\mF}_t$.

\ENDFOR

\ENSURE Retained parameters $\vtheta_T$ and consolidated importance
$\mLambda_T$.

\end{algorithmic}
\end{algorithm}

\subsubsection{Approximation Error Analysis}

\minisection{Finite-Difference Approximation}
\label{app:fs_fd_accuracy}
The estimator in
Eq.~\ref{eq:squared_fisher_gradient_estimator}
approximates the exact squared-gradient derivative at a fixed
parameter point. By the fundamental theorem of calculus,
\begin{equation}
\widehat{\vh}(\vu)
=
2
\int_0^1
\mH
\bigl(
\vu+\tau\rho\mathbf{s}(\vu)
\bigr)
\vg(\vu)
\,\mathrm d\tau.
\label{eq:app_fs_estimator_integral}
\end{equation}
Hence, continuity of the Hessian implies
% \begin{equation}
$
\widehat{\vh}(\vu)
\rightarrow
2\mH(\vu)\vg(\vu)
\quad
\text{as }
\rho\rightarrow0.
\label{eq:app_fs_estimator_limit}
$
% \end{equation}

If the Hessian is $L_H$-Lipschitz along the probe segment, then
\begin{equation}
\left\|
\widehat{\vh}(\vu)
-
2\mH(\vu)\vg(\vu)
\right\|_2
\leq
L_H |\rho|
\frac{\|\vg(\vu)\|_2^2}
{\|\vg(\vu)\|_2+\xi}.
\label{eq:app_fs_finite_difference_error}
\end{equation}
Thus, the finite-difference estimator has first-order truncation
error with respect to the probe scale $\rho$.
% This result concerns derivative estimation at a fixed parameter
% point and is separate from the approximation used to locate the
% neighborhood maximum.
% Unless otherwise specified, we use $\rho=1$ in the experiments.

\minisection{Single-Step Inner-Search Approximation}
\label{app:fs_search_accuracy}

The neighborhood point $\vz$ is obtained from a single first-order
ascent step rather than by solving the inner maximization exactly.
For a perturbation $\vepsilon$ around $\vtheta$,
\begin{equation}
\|\vg(\vtheta+\vepsilon)\|_2^2
=
\|\vg(\vtheta)\|_2^2
+
\left[
2\mH(\vtheta)\vg(\vtheta)
\right]^\top
\vepsilon
+
o(\|\vepsilon\|_2).
\label{eq:app_fs_inner_linearization}
\end{equation}
Therefore, the normalized direction of
$2\mH(\vtheta)\vg(\vtheta)$ maximizes the linearized sensitivity
objective over the radius-$r$ ball.
The practical solver replaces this exact sensitivity gradient with
its finite-difference estimate and uses Eq.~\ref{eq:squared_fisher_numerical_update}.
% \begin{equation}
% \widehat{\vepsilon}(\vtheta)
% =
% r
% \frac{\widehat{\vh}(\vtheta)}
% {\|\widehat{\vh}(\vtheta)\|_2+\xi},
% \qquad
% \vz
% =
% \vtheta+\widehat{\vepsilon}(\vtheta).
% \label{eq:app_fs_practical_inner_step}
% \end{equation}

Let
% \begin{equation}
$
\vepsilon^\star
\in
\arg\max_{\|\vepsilon\|_2\leq r}
\|\vg(\vtheta+\vepsilon)\|_2^2
\label{eq:app_fs_exact_inner_maximizer}
% \end{equation}
$
denote an exact inner maximizer.
The practical point $\vz$ need not coincide with
$\vtheta+\vepsilon^\star$ because it is obtained from a single
local ascent step.
Consequently, the practical solver introduces two distinct
approximations: finite-difference estimation of the sensitivity
gradient and single-step approximation of the neighborhood
maximization.
The probe scale $\rho$ affects the former, whereas the neighborhood
radius $r$ controls the search region of the latter.

\section{Supplementary Experiments}
\label{app:supp_experiments}

\subsection{Experimental Configurations}
\label{app:experimental_configurations}

We organize the experimental setup into three groups:
(i) standard vision benchmarks,
(ii) fine-grained and heterogeneous vision benchmarks, and
(iii) continual text classification.
% Within each group, we distinguish the shared benchmark protocol from
% dataset-specific and method-specific settings.
For published baselines, we follow the configurations reported by the
original papers or released implementations whenever available.
For EWC and FS-EWC, we report the configurations associated
with the corresponding experimental runs.
% Accordingly, methods within the same evaluation group share the main
% benchmark protocol, but are not assumed to use identical optimization
% hyperparameters when their original implementations require different
% settings.

\paragraph{Standard Vision LoRA Benchmarks}
\label{app:standard_vision_settings}

We evaluate on CIFAR-100, DomainNet, ImageNet-R, and ImageNet-A
following the EWC-LoRA~\citep{zheng2026ewclora}.
CIFAR-100, ImageNet-R, and ImageNet-A each contain $10$ sequential
tasks, whereas DomainNet contains $5$ tasks.

We use the same ViT-B/16 backbone, pretrained initialization,
LoRA placement, and dataset-specific ranks as the reference setting,
with ranks $10$, $30$, $10$, and $10$, respectively.
% Unless otherwise stated, the training length and baseline-specific
% hyperparameters follow the same reference.
The reported Vanilla LoRA, InfLoRA~\citep{liangInfLoRAInterferenceFreeLowRank2024a}, SD-LoRA~\citep{wu2025sdlorascalabledecoupledlowrank}, CL-LoRA~\citep{CLLORA}, and BiLoRA~\citep{ICLR2026_9c84feb7} results are taken directly from
\citet{zheng2026ewclora}.
We refer readers to that work for their optimizer,
learning-rate, classifier, and method-specific regularization settings.

FS-EWCLoRA follows the same task construction, backbone,
LoRA placement, and dataset-specific ranks as the above benchmark.
We use SGD with batch size 128 and cosine learning-rate scheduling.
The shaping radius is $r=0.05$ across datasets.
The probe radius is $\rho=0.1$ on CIFAR-100,
and $\rho=1$ on ImageNet-A, ImageNet-R and DomainNet.
The damping coefficient is $0.2$.
Checkpoints are collected every $5$, $5$, $20$, and $20$ epochs, for
CIFAR-100, DomainNet, ImageNet-R, and ImageNet-A, respectively.

% \paragraph{Full-model EWC and FS-EWC.}
We additionally evaluate EWC and FS-EWC under full-model fine-tuning
on the same four benchmarks.
All runs use ViT-B/16, SGD, learning rate $0.01$, batch size 128,
cosine scheduling, zero momentum, and zero weight decay.
The numbers of epochs per task are $20$, $5$, $30$, and $20$
for CIFAR-100, DomainNet, ImageNet-R, and ImageNet-A, respectively.
The EWC and FS-EWC regularization coefficients are
$4\times10^4$, $1\times10^5$, $8\times10^4$, and $2\times10^5$
for CIFAR-100, DomainNet, ImageNet-R, and ImageNet-A, respectively.
% while FS-EWC uses
% $4\times10^4$, $1\times10^5$, $8\times10^4$, and $2\times10^3$.
For FS-EWC, the shaping radius $r$ / finite-difference probe scale
$\rho$ are set to
$0.005/0.10$, $0.05/0.20$, $0.05/0.20$, and $0.01/0.20$,
% For FS-EWC, the shaping radius $r$ are set to
% $0.005$, $0.05$, $0.05$, and $0.01$,
with retained-checkpoint intervals of $5$, $1$, $10$, and $20$
epochs, respectively.
All full-model FS-EWC runs use damping coefficient $0.2$.
Fisher statistics are estimated from at most 100 current-task batches
and exclude the classification head.

% The EWC regularization coefficients are
% $4\times10^4$, $1\times10^5$, $8\times10^4$, and $2\times10^5$,
% while FS-EWC uses
% $4\times10^4$, $1\times10^5$, $8\times10^4$, and $2\times10^3$,
% respectively.
% Fisher statistics are estimated from at most 100 current-task batches
% and exclude the classification head.

% For FS-EWC, the shaping radius $r$ / normalized probe radius $\delta$ are
% $0.005/0.01$, $0.05/0.20$, $0.05/0.20$, and $0.01/0.20$.
% The retained-checkpoint intervals are $5$, $1$, $10$, and $20$ epochs.
% All full-model FS-EWC runs use damping coefficient $0.2$.

\paragraph{Fine-Grained and Heterogeneous Vision Benchmarks}
\label{app:finegrained_settings}

We further evaluate LoRA-based continual learning on Aircraft, Cars,
CUB-200, and Flowers, together with the heterogeneous
5Datasets benchmark.
Aircraft, Cars, CUB-200, and Flowers are divided into $10$ sequential
tasks, while Pets contains $9$ tasks.
For 5Datasets, we follow the task order
CUB-200 $\rightarrow$ Aircraft $\rightarrow$ Cars
$\rightarrow$ Flowers $\rightarrow$ Pets.
The detailed task partitions are summarized in
Table~\ref{tab:finegrained_task_splits}. 
\begin{table}[ht]
\centering
\captionsetup{font=footnotesize, labelfont=bf}
\caption{
Task partitions for the fine-grained and heterogeneous vision
benchmarks.
}
\label{tab:finegrained_task_splits}
\scriptsize
\setlength{\tabcolsep}{5pt}
\renewcommand{\arraystretch}{1.05}
\begin{tabular}{lccc}
\toprule
Dataset
& Classes
& Tasks
& Task partition
\\
\midrule
Aircraft
& 100 & 10 & $10\times10$ \\
Cars
& 196 & 10 & $16 + 9\times20$ \\
CUB-200
& 200 & 10 & $20\times10$ \\
Flowers
& 102 & 10 & $12 + 9\times10$ \\
% Pets
% & 37 & 9 & $5 + 8\times4$ \\
5Datasets
& 635 & 5 & $200/100/196/102/37$ \\
\bottomrule
\end{tabular}
\end{table}

All methods use ViT-B/16 initialized from an ImageNet-1K
fine-tuned checkpoint.
LoRA is applied to the query and value projections with rank $16$.
We use SGD with batch size $128$, zero momentum, zero weight decay,
and cosine learning-rate scheduling.
% All experiments are replay-free and use seed $0$.
For the individual fine-grained datasets, models are trained for
$40$ epochs per task.
The learning rates are $0.05$ for Aircraft and Cars,
$0.01$ for CUB-200, and $0.0025$ for Flowers.
For 5Datasets, we use learning rate $0.02$.

% \paragraph{Training configuration.}

% \paragraph{EWC and FS-EWC settings.}
For both EWC-LoRA and FS-EWC-LoRA, we use the same LoRA rank,
optimizer, learning rate, and training schedule described above,
with the EWC regularization coefficient fixed to
$\lambda=2000$ on all benchmarks.
FS-EWCLoRA additionally uses shaping radius $r=0.05$
and
finite-difference probe scale $\rho=0.2$.
The current runs retain only the task-end model, so no
multi-checkpoint averaging is applied in this experiment.

\paragraph{Continual Text Classification}
\label{app:nlp_settings}

We evaluate on four text-classification datasets from the continual
learning benchmark of~\citet{NIPS2015_250cf8b5}: AG News,
Amazon Reviews, DBpedia, and Yahoo Answers.
We consider three task orders to evaluate robustness to different
task sequences as shown in Table \ref{tab:nlp_task_orders}.

\begin{table}[bht]
\centering
\captionsetup{font=footnotesize, labelfont=bf}
\caption{Task orders for continual text classification.}
\label{tab:nlp_task_orders}
\scriptsize
\setlength{\tabcolsep}{5pt}
\renewcommand{\arraystretch}{1.05}
\begin{tabular}{cl}
\toprule
Order & Task sequence \\
\midrule
1 &
DBpedia $\rightarrow$ Amazon $\rightarrow$ Yahoo
$\rightarrow$ AG News
\\
2 &
DBpedia $\rightarrow$ Amazon $\rightarrow$ AG News
$\rightarrow$ Yahoo
\\
3 &
Yahoo $\rightarrow$ Amazon $\rightarrow$ AG News
$\rightarrow$ DBpedia
\\
\bottomrule
\end{tabular}
\end{table}

We use Llama-3.2-1B and Llama-3.2-3B with LoRA applied to the
query and value projections.
All experiments use LoRA rank $8$, scaling factor $32$, dropout
$0.1$, BF16 precision, and one training epoch per task.
The first task uses learning rate $10^{-3}$ and subsequent tasks use
$10^{-4}$ with a constant schedule and no warmup.
The maximum input length is $512$, the target/generation length is
$50$, gradient clipping is $1.0$, and the effective batch size is
$64$ following OLoRA~\citep{wang2023orthogonal}.
AdamW is used for optimization.

% \paragraph{Training configuration.}

% \paragraph{EWC and FS-EWC-LoRA settings.}
For EWC-LoRA, the EWC coefficient is $\lambda=0.5$ for
Llama-3.2-1B and $\lambda=1$ for Llama-3.2-3B.
FS-EWC-LoRA uses $\lambda=1$ for both model sizes,
together with shaping radius $r=0.02$ and finite-difference probe
scale $\delta=0.2$.
The remaining shaping coefficients are fixed across all task orders.
Following EWCLoRA, we also adopt TreeLoRA~\citep{qian2026treeloraefficientcontinuallearning} for comparison.
% \section{Supplementary Experiments}
% \label{app:experiments}

% \subsection{Detailed Experimental Protocol}
% \label{app:exp_protocol}

% % \paragraph{Statistical reporting.}
% % The following is the reporting specification for the final tables.
% % Use matched seeds and task orders for paired comparisons.
% % State whether reported dispersion is over seeds, orders, or both.
% % For the three NLP orders, first compute each metric per run, summarize
% % seed variation within an order, and only then summarize across orders.
% % Report all three order-specific effects; three orders are not a
% % substitute for repeated random seeds under one order.
% % Do not count tasks or directions from a single run as independent
% % training replications. Predeclare the validation and tuning protocol;
% % do not select configurations using historical test or diagnostic data.

% \paragraph{Metric identity.}
% Task-averaged FAA/AAA, pooled-example accuracy, old-class accuracy,
% and task-local accuracy must retain distinct names.
% An expanding all-seen-class softmax changes its candidate set over time.
% Finite-displacement loss probes therefore keep their loss function
% and candidate set fixed, while class-incremental accuracy is evaluated
% separately over all classes seen at the evaluation stage.
% Normalized stability/plasticity, when retained from existing runs,
% use their recorded definitions and reference models: .
% They are not renamed BWT, LA, or forward transfer.

\subsection{Additional Experimental Result}

\paragraph{Stage-wise performance on fine-grained and heterogeneous benchmarks.}
\label{app:additional_results}

\begin{figure}[ht]
    \centering
    \includegraphics[width=\linewidth]
    {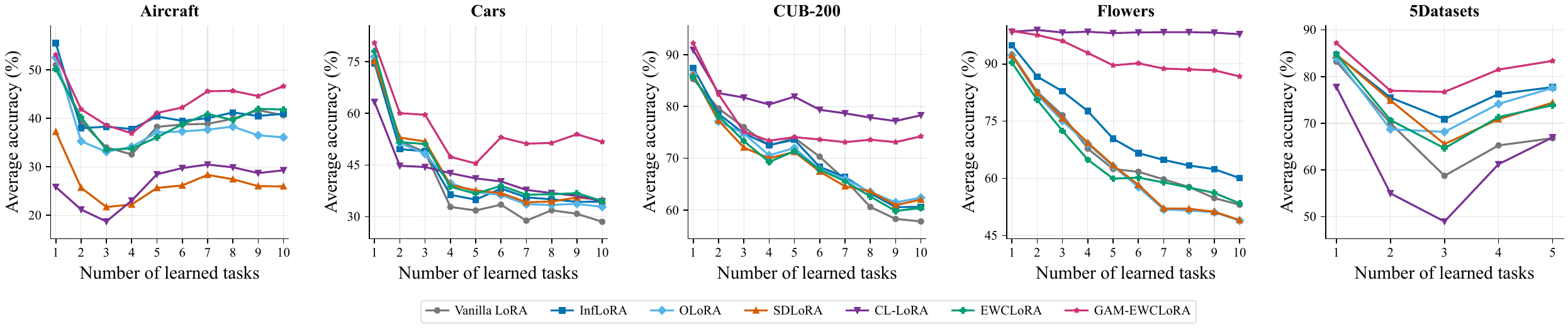}
    \caption{
    Stage-wise average-accuracy trajectories on the fine-grained and
    heterogeneous vision benchmarks reported in
    Table~\ref{tab:exp_fine}.
    % The x-axis denotes the number of learned tasks, and the y-axis
    % reports the average accuracy over all tasks observed up to each stage.
    }
    \label{fig:finegrained_stagewise}
\end{figure}

Figure~\ref{fig:finegrained_stagewise} complements the aggregate
results in Table~\ref{tab:exp_fine} by showing how average accuracy
evolves as new tasks are introduced.
FS-EWCLoRA exhibits particularly strong late-stage performance on
Aircraft, Cars, and 5Datasets.
On CUB-200 and Flowers, although CL-LoRA remains stronger,
FS-EWCLoRA consistently improves over its matched EWCLoRA baseline.
Overall, the trajectories indicate that the improvements are generally
accumulated during sequential learning rather than arising only at the
final stage.

\paragraph{Stage-wise performance on continual NLP benchmark.}

\begin{figure}[bht]
    \centering
    \includegraphics[width=1\linewidth]{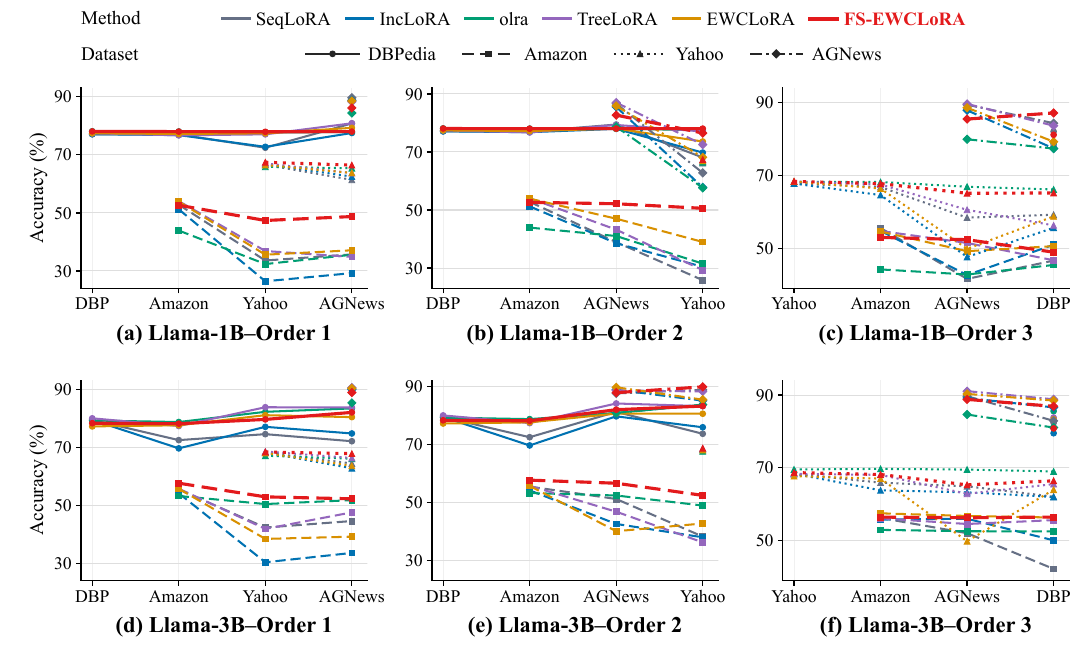}
    \caption{Task-wise accuracy trajectories for Llama-3.2-1B and Llama-3.2-3B
across three continual text-classification task orders.
Each curve starts when the corresponding dataset is learned and tracks
its accuracy after subsequent tasks.}
    \label{fig:llama_taskwise_combined_v6}
\end{figure}

Fig.~\ref{fig:llama_taskwise_combined_v6} complements
Table~\ref{tab:exp_nlp} by showing how individual task accuracies
evolve under subsequent learning.
Across most orders, FS-EWCLoRA better preserves previously learned
tasks, particularly those exposed to multiple later updates, while
the gains are not necessarily uniform across every dataset.
The main exception is Llama-3.2-3B under Order~3, where weaker final
performance on some tasks offsets improved retention on others and
produces a slightly lower FAA than EWCLoRA.
Overall, the trajectories support the more consistent performance of
FS-EWCLoRA across the evaluated task orders.

\subsection{Additional Two-Task Solution-Shaping Diagnostics}
\label{app:two_task_diagnostics}

Table~\ref{tab:two_task_parameterization_diagnostics} shows a
consistent pattern across both parameterizations.
Under the same $\mathsf C+\mathsf P$ objective on Task~2, shaping
the inherited Task~1 solution increases its retained accuracy from
$87.95$ to $91.82$ under LoRA and from $85.53$ to $87.89$ under
full-parameter fine-tuning.
The complete FS-EWC configuration also yields higher final Task~1
accuracy than EWC in both settings, while improving Task~2 accuracy
under LoRA and leaving it nearly unchanged under full fine-tuning.
For LoRA, the more negative BWT of FS-EWC reflects its higher
Task~1 starting accuracy rather than lower final retention.
These results reproduce the main mechanism across parameterizations:
subsequent retention depends on the solution passed forward as well
as on the protection subsequently applied to it.

\begin{table*}[h]
\centering
\captionsetup{font=footnotesize,labelfont=bf}
\caption{
Two-task solution-shaping diagnostics under LoRA adaptation and
full-parameter fine-tuning on ImageNet-R with ViT-B/16.
$\mathsf C$, $\mathsf F$, and $\mathsf P$ denote current-task fitting,
Fisher shaping, and historical EWC protection, respectively.
$\mathcal F_1$ denotes the Task-1 Fisher statistic reported under each
parameterization, and $d_2$ denotes the Task-2 parameter displacement
in the corresponding trainable parameter space.
% LoRA results are means over three seeds, with standard deviations
% omitted for compactness.
All variants use endpoint references without trajectory averaging.
}
\label{tab:two_task_parameterization_diagnostics}

\scriptsize
\setlength{\tabcolsep}{2.5pt}
\renewcommand{\arraystretch}{1.08}

\begin{tabular*}{\textwidth}{
@{\extracolsep{\fill}}llcccccccc@{}
}
\toprule
\multirow{2}{*}{Model}
& \multirow{2}{*}{Variant}
& \multicolumn{2}{c}{Training objective}
& \multicolumn{2}{c}{Task 1 endpoint}
& \multicolumn{3}{c}{Task 2 endpoint}
& \multicolumn{1}{c}{Task 2 update}
\\
\cmidrule(lr){3-4}
\cmidrule(lr){5-6}
\cmidrule(lr){7-9}
\cmidrule(lr){10-10}

&
& Task 1
& Task 2
& $a_{11}\uparrow$
& $\mathcal F_1$
& $a_{21}\uparrow$
& $a_{22}\uparrow$
& BWT$\uparrow$
& $d_2$
\\
\midrule

% =========================================================
% LoRA
% =========================================================

\multirow{6}{*}{LoRA}
& Vanilla
& $\mathsf C$
& $\mathsf C$
& 90.73
& 19.885
& 84.85
& 89.59
& -5.87
& 1.560
\\

& Shaping only
& $\mathsf C+\mathsf F$
& $\mathsf C+\mathsf F$
& 93.34
& 10.946
& 84.80
& 93.81
& -8.55
& 3.131
\\

& EWC
& $\mathsf C$
& $\mathsf C+\mathsf P$
& 90.73
& 19.885
& 87.95
& 83.45
& -2.78
& 0.0106
\\

& FS-EWC
& $\mathsf C+\mathsf F$
& $\mathsf C+\mathsf F+\mathsf P$
& 93.34
& 10.946
& 89.10
& 88.33
& -4.24
& 0.0204
\\

& Task-1 shaping
& $\mathsf C+\mathsf F$
& $\mathsf C+\mathsf P$
& 93.34
& 10.946
& 91.82
& 82.57
& -1.52
& 0.392
\\

& Task-2 shaping
& $\mathsf C$
& $\mathsf C+\mathsf F+\mathsf P$
& 90.73
& 19.885
& 83.07
& 89.21
& -7.66
& 0.129
\\

\midrule

% =========================================================
% Full-parameter fine-tuning
% =========================================================

\multirow{6}{*}{Full FT}
& Vanilla
& $\mathsf C$
& $\mathsf C$
& 92.61
& 869.80
& 78.46
& 92.47
& -14.15
& 4.941
\\

& Shaping only
& $\mathsf C+\mathsf F$
& $\mathsf C+\mathsf F$
& 93.24
& 858.59
& 81.92
& 93.44
& -11.32
& 5.131
\\

& EWC
& $\mathsf C$
& $\mathsf C+\mathsf P$
& 92.61
& 869.80
& 85.53
& 90.66
& -7.08
& 4.747
\\

& FS-EWC
& $\mathsf C+\mathsf F$
& $\mathsf C+\mathsf F+\mathsf P$
& 93.24
& 858.59
& 90.25
& 90.79
& -2.99
& 4.673
\\

& Task-1 shaping
& $\mathsf C+\mathsf F$
& $\mathsf C+\mathsf P$
& 93.24
& 858.59
& 87.89
& 87.31
& -5.35
& 4.774
\\

& Task-2 shaping
& $\mathsf C$
& $\mathsf C+\mathsf F+\mathsf P$
& 92.61
& 869.80
& 82.08
& 93.86
& -10.53
& 4.595
\\

\bottomrule
\end{tabular*}
\end{table*}

\subsection{Complementarity Study: Experimental Details and Sensitivity to $\lambda$}
\label{app:complementarity_details}

% Requires: booktabs, multirow, caption

We further evaluate the Fisher-complementarity study over three
random seeds.
For all Fisher computations, we keep the historical regularization
coefficient fixed across methods and seeds, using
$\lambda=2\times10^6$ for the diagonal representation and
$\lambda=2\times10^5$ for the Kronecker-factored representation.
These coefficients are also shared across the batched empirical,
per-example empirical, and model-sampled Fisher variants.
Unless otherwise specified, all remaining training and
retained-state construction settings follow
Table~\ref{tab:complementarity}.

\begin{table*}[ht]
\centering
\captionsetup{font=footnotesize, labelfont=bf}
\caption{
Multi-seed results for the Fisher-complementarity study in
Table~\ref{tab:complementarity}.
FAA and AAA are reported as mean $\pm$ sample standard deviation
(\%) over seeds 0, 1, and 2.
}
\label{tab:complementarity_multiseed}

\footnotesize
\setlength{\tabcolsep}{3.2pt}
\renewcommand{\arraystretch}{1.05}

\begin{tabular*}{\textwidth}{
@{\extracolsep{\fill}}llcccccc@{}
}
\toprule

\multirow{2}{*}{Method}
& \multirow{2}{*}{Metric}
& \multicolumn{3}{c}{Diagonal}
& \multicolumn{3}{c}{Kronecker-factored} \\
\cmidrule(lr){3-5}
\cmidrule(lr){6-8}

&
& Batched
& Per-example
& Model-sampled
& Batched
& Per-example
& Model-sampled \\

\midrule

\multirow{2}{*}{EWC}
& FAA
& \ExpRes{51.96}{1.15}
& \ExpRes{50.60}{1.07}
& \ExpRes{50.76}{1.14}
& \ExpRes{53.74}{0.47}
& \ExpRes{52.75}{0.78}
& \ExpRes{52.71}{1.17}
\\

& AAA
& \ExpRes{66.30}{0.53}
& \ExpRes{66.60}{0.97}
& \ExpRes{66.60}{1.09}
& \ExpRes{67.37}{0.18}
& \ExpRes{67.55}{0.94}
& \ExpRes{67.33}{0.89}
\\

\addlinespace[2pt]

\multirow{2}{*}{FS-EWC}
& FAA
& \ExpRes{59.29}{0.55}
& \ExpRes{59.67}{1.07}
& \ExpRes{59.62}{0.88}
& \ExpRes{59.82}{0.91}
& \ExpRes{59.95}{0.93}
& \ExpRes{59.62}{0.82}
\\

& AAA
& \ExpRes{70.69}{0.67}
& \ExpRes{70.88}{0.50}
& \ExpRes{70.86}{0.51}
& \ExpRes{71.00}{0.78}
& \ExpRes{71.16}{0.47}
& \ExpRes{71.12}{0.46}
\\

\bottomrule
\end{tabular*}

\par\vspace{2pt}
% \begin{minipage}{\textwidth}
% \footnotesize
% \textit{Note.}
% Batched uses batch-gradient, Per-example uses
% individual empirical gradients, and Model-sampled uses one
% model-sampled label per example.
% We use $\lambda=2\times10^6$ for the diagonal representation and
% $\lambda=2\times10^5$ for the Kronecker-factored representation,
% with the same value shared by EWC and FS-EWC across Fisher
% computations and random seeds.
% \end{minipage}

\end{table*}

\paragraph{Multi-seed consistency.}
Table~\ref{tab:complementarity_multiseed} shows that FS-EWC
improves over the matched EWC baseline under all six combinations
of Fisher computation and matrix representation.
The improvements range from 6.08 to 9.07 percentage points in FAA
and from 3.61 to 4.39 points in AAA.
Kronecker-factored Fisher generally provides modest gains over the
diagonal representation, particularly for EWC, while the advantage
of FS-EWC persists under both representations and all three Fisher
computations.
These results confirm that the complementarity observed in
Table~\ref{tab:complementarity} is consistent across random seeds.

\paragraph{Sensitivity to historical regularization strength.}
Table~\ref{tab:lambda_sensitivity_complementarity} further examines
the historical regularization coefficient $\lambda$ under batched
empirical Fisher.
Across both diagonal and Kronecker-factored representations,
FS-EWC remains above the corresponding EWC baseline at every
numerically valid matched value of $\lambda$.
Its performance also varies relatively little over the evaluated
range, whereas EWC is more sensitive to the strength of historical
regularization.
This indicates that the gain from solution shaping is not confined
to a narrowly selected protection coefficient.

% The coefficients used in the main complementarity study were fixed
% from the original exploratory search:
% $\lambda=2\times10^6$ for the diagonal representation and
% $\lambda=2\times10^5$ for the Kronecker-factored representation.
% These values were subsequently transferred unchanged to the
% per-example, model-sampled, and multi-seed experiments.
% They were not reselected after updating the FAA/AAA computation.
% Under the current metric, the Kronecker-factored FS-EWC run reaches
% a higher value at $\lambda=2\times10^6$, whereas the corresponding
% EWC run becomes numerically unstable.
% We therefore retain the previously fixed shared coefficient rather
% than retrospectively selecting different values for the two methods.

\begin{table*}[h]
\centering
\captionsetup{font=footnotesize, labelfont=bf}
\caption{
Sensitivity to the historical regularization coefficient $\lambda$
under batched empirical Fisher.
Each entry reports FAA / AAA (\%) for seed 2.
% The analysis uses the same experimental setting as
% Table~\ref{tab:complementarity}.
% $\dagger$ marks the value of $\lambda$ retained for the corresponding
% Fisher representation in the main complementarity study.
% All results in this table use seed 2.
}
\label{tab:lambda_sensitivity_complementarity}

\footnotesize
\setlength{\tabcolsep}{3.2pt}
\renewcommand{\arraystretch}{1.05}

\begin{tabular*}{\textwidth}{
@{\extracolsep{\fill}}llcccccc@{}
}
\toprule
\multirow{2}{*}{Representation}
& \multirow{2}{*}{Method}
& \multicolumn{6}{c}{Historical regularization coefficient $\lambda$} \\
\cmidrule(lr){3-8}
&
& $0$
& $2\times10^2$
& $2\times10^3$
& $2\times10^4$
& $2\times10^5$
& $2\times10^6$ \\
\midrule

\multirow{2}{*}{Diagonal}
& EWC
& 49.51 / 64.95
& 49.24 / 64.66
& 49.37 / 64.66
& 49.70 / 65.01
& 50.56 / 65.55
& 52.07 / 66.38
\\
& FS-EWC
& 58.53 / 69.82
& 58.79 / 69.65
& 58.39 / 69.75
& 58.59 / 69.66
& 58.39 / 69.73
& 58.66 / 69.94
\\

\midrule

\multirow{2}{*}{\shortstack{Kronecker-\\factored}}
& EWC
& 49.31 / 64.92
& 49.31 / 64.89
& 49.70 / 65.15
& 50.23 / 66.15
& 53.32 / 67.33
& NaN
\\
& FS-EWC
& 58.85 / 69.80
& 58.39 / 69.78
& 58.66 / 69.80
& 58.46 / 69.81
& 58.79 / 70.14
& 58.92 / 70.57
\\

\bottomrule
\end{tabular*}

\vspace{2pt}
\begin{minipage}{\textwidth}
\footnotesize
The Kronecker-factored EWC run at
$\lambda=2\times10^6$ encountered numerical instability and produced
NaN values.
\end{minipage}
\end{table*}

\subsection{Retained-State Construction Details}
\label{app:retained_state_details}

Table~\ref{tab:retained_state_multiseed} shows that trajectory
averaging achieves the highest AAA for all three seeds, whereas
denser trajectory sampling provides no further improvement.
All multi-checkpoint constructions yield less negative BWT than the
endpoint reference, although endpoint-plus-initial achieves the least
negative BWT at the cost of substantially lower FAA and AAA.
Including the initialization provides no additional accuracy benefit
for either endpoint or trajectory averaging.
Overall, these results indicate that learned states along the training
trajectory provide useful information for retained-state construction,
but neither denser sampling nor simply averaging more states is
inherently beneficial.

\begin{table*}[h]
\centering
\captionsetup{font=footnotesize,labelfont=bf}
\caption{
Multi-seed results for retained-state construction on ImageNet-A.
Each retained state is constructed by uniformly averaging the LoRA
factors at checkpoints $\mathcal{K}_t$, with $\omega=1$; the
current-task Fisher is estimated after merging.
Checkpoint $0$ denotes the state before the first optimizer update
of each task, while checkpoint $50$ denotes the terminal iterate.
}
\label{tab:retained_state_multiseed}

\small
\setlength{\tabcolsep}{5.0pt}
\renewcommand{\arraystretch}{1.08}

\begin{tabular}{@{}llcrrrr@{}}
\toprule
Retained state
& Checkpoints $\mathcal{K}_t$
& Metric
& Seed 0
& Seed 1
& Seed 2
& Mean $\pm$ SD \\
\midrule

\multirow{3}{*}{Endpoint}
& \multirow{3}{*}{$\{50\}$}
& FAA
& 58.20
& 59.97
& 59.51
% & $59.23 \pm 0.92$ \\
& \ExpRes{59.23}{0.92} \\
&
& AAA
& 70.48
& 71.00
& 70.41
% & $70.63 \pm 0.33$ \\
& \ExpRes{70.63}{0.33} \\
&
& BWT
& $-7.29$
& $-8.66$
& $-8.29$
% & $-8.08 \pm 0.71$ \\
& \ExpRes{-8.08}{0.71} \\

\midrule

\multirow{3}{*}{Late-stage}
& \multirow{3}{*}{$\{30,40,50\}$}
& FAA
& 58.53
& 59.84
& 59.91
% & $\mathbf{59.43 \pm 0.78}$ \\
& \textbf{\ExpRes{59.43}{0.78}} \\
&
& AAA
& 70.51
& 70.98
& 70.35
% & $70.61 \pm 0.33$ \\
& \ExpRes{70.61}{0.33} \\
&
& BWT
& $-7.01$
& $-8.35$
& $-7.68$
% & $-7.68 \pm 0.67$ \\
& \ExpRes{-7.68}{0.67} \\

\midrule

\multirow{3}{*}{Trajectory}
& \multirow{3}{*}{$\{10,20,30,40,50\}$}
& FAA
& 58.59
& 59.78
& 59.64
% & $59.34 \pm 0.65$ \\
& \ExpRes{59.34}{0.65} \\
&
& AAA
& 70.63
& 71.29
& 70.61
% & $\mathbf{70.84 \pm 0.39}$ \\
& \textbf{\ExpRes{70.84}{0.39}} \\
&
& BWT
& $-6.89$
& $-7.70$
& $-7.81$
% & $-7.47 \pm 0.50$ \\
& \ExpRes{-7.47}{0.50} \\

\midrule

\multirow{3}{*}{Dense trajectory}
& \multirow{3}{*}{$\{5,10,\ldots,50\}$}
& FAA
& 58.39
& 59.64
& 59.91
% & $59.31 \pm 0.81$ \\
& \ExpRes{59.31}{0.81} \\
&
& AAA
& 70.55
& 71.15
& 70.52
% & $70.74 \pm 0.35$ \\
& \ExpRes{70.74}{0.35} \\
&
& BWT
& $-7.00$
& $-8.01$
& $-7.32$
% & $-7.44 \pm 0.52$ \\
& \ExpRes{-7.44}{0.52} \\

\midrule

\multirow{3}{*}{Endpoint + initial}
& \multirow{3}{*}{$\{0,50\}$}
& FAA
& 56.88
& 58.92
& 59.51
% & $58.44 \pm 1.38$ \\
& \ExpRes{58.44}{1.38} \\
&
& AAA
& 68.87
& 69.72
& 69.56
% & $69.38 \pm 0.45$ \\
& \ExpRes{69.38}{0.45} \\
&
& BWT
& $-6.52$
& $-6.34$
& $-6.44$
% & $\mathbf{-6.44 \pm 0.09}$ \\
& \textbf{\ExpRes{-6.44}{0.09}} \\

\midrule

\multirow{3}{*}{Trajectory + initial}
& \multirow{3}{*}{$\{0,10,20,30,40,50\}$}
& FAA
& 57.93
& 59.38
& 59.71
% & $59.01 \pm 0.95$ \\
& \ExpRes{59.01}{0.95} \\
&
& AAA
& 70.32
& 70.94
& 70.49
% & $70.59 \pm 0.32$ \\
& \ExpRes{70.59}{0.32} \\
&
& BWT
& $-6.84$
& $-7.58$
& $-7.57$
% & $-7.33 \pm 0.43$ \\
& \ExpRes{-7.33}{0.43} \\

\bottomrule
\end{tabular}
\end{table*}

\end{document}